\documentclass[sigconf]{acmart}
\setcopyright{none}
\renewcommand\footnotetextcopyrightpermission[1]{} 

\usepackage{graphicx}
\usepackage{microtype}
\usepackage{graphicx}
\usepackage{subfigure}
\usepackage{lipsum}
\usepackage{booktabs} 
\usepackage{hyperref}
\usepackage{stackengine}
\usepackage{url}
\usepackage{courier}  
\usepackage{latexsym}
\usepackage{latexsym}
\usepackage{wrapfig, blindtext}
\usepackage{multirow}
\usepackage{amsmath}
\usepackage{float}
\usepackage{booktabs}
\usepackage{siunitx}
\usepackage{hyperref}
\usepackage{mathtools}
\usepackage{capt-of}
\usepackage{algorithm}
\usepackage{algpseudocode}

\DeclarePairedDelimiter\norm{\lVert}{\rVert}
\DeclarePairedDelimiter\abs{\lvert}{\rvert}
\DeclareMathOperator{\sign}{sign}

\AtBeginDocument{%
  \providecommand\BibTeX{{%
    \normalfont B\kern-0.5em{\scshape i\kern-0.25em b}\kern-0.8em\TeX}}}

\begin{document}

\title{Knowledge-Guided Exploration in Deep Reinforcement Learning}




\author{Sahisnu Mazumder$^{1~*}$,
Bing Liu$^1$, 
Shuai Wang$^1$  
}\authornote{Work done while author was affiliated to University of Illinois at Chicago, USA}
\author{
Yingxuan Zhu$^{2~\dagger}$, 
Xiaotian Yin$^{2~\dagger}$, 
Lifeng Liu$^{2~\dagger}$, 
Jian Li$^{2}$ 
}
\authornote{Work done while author was affiliated to FutureWei Technologies, MA, USA}
\vspace{0.1cm}
\affiliation{%
  \institution{$^1$ Department of Computer Science, University of Illinois at Chicago, USA }
}
\email{sahisnumazumder@gmail.com, liub@cs.uic.edu, shuaiwanghk@gmail.com}
\vspace{0.1cm}
\affiliation{%
  \institution{$^2$ FutureWei Technologies, 1 Broadway, Cambridge, USA}
}
\email{{yingxuan.zhu,xiaotian.yin,lifeng.liu1,jian.li1}@huawei.com}
 
\renewcommand{\shortauthors}{S. Mazumder et al.}

\begin{abstract}
This paper proposes a new method to drastically speed up deep reinforcement learning (deep RL) training for problems that have the property of \textit{state-action permissibility} (SAP). Two types of permissibility are defined under SAP. The first type says that after an action $a_t$ is performed in a state $s_t$ and the agent have reached the new state $s_{t+1}$, the agent can decide whether $a_t$ is \textit{permissible} or \textit{not permissible} in $s_t$. The second type says that even without performing $a_t$ in $s_t$, the agent can already decide whether $a_t$ is permissible or not in $s_t$. An action is not permissible in a state if the action can never lead to an optimal solution and thus should not be tried (over and over again). We incorporate the proposed SAP property and encode action permissibility knowlegde into two state-of-the-art deep RL algorithms to guide their state-action exploration together with a \textit{virtual stopping} strategy. Results show that the SAP-based guidance can markedly speed up RL training.
\end{abstract}



\keywords{Reinforcement Learning, Action Exploration, RL speedup.}


\maketitle

\section{Introduction}
\label{intro}
Exploration-exploitation trade-off~\cite{sutton2017reinforcement} has been a persistent problem that makes RL slow. A major reason behind this problem is that unlike humans, a model-free RL agent has no or limited understanding of the dynamics/properties of the underlying task or environment and thus, lacks the \textit{prior knowledge} to guide its exploration. Recent studies \cite{Tsividis2017,Dubey2018} have shown the importance of such ``\textit{prior knowledge}" that enables humans to learn a task (e.g., playing a video game) much faster. We believe a major attempt should be made to study \textbf{knowledge-guided RL}, i.e., encoding external knowledge to guide RL agents in exploration, to make RL efficient for practical applications. 

In most applications, some properties or knowledge of the tasks can be exploited to drastically speed up RL training. This paper identifies such a property, called \textit{state-action permissibility} (SAP), which can markedly reduce RL training time for the class of problems with the property. In our work, we develop a framework for incorporating SAP-based guidance into Deep RL for problems with one-dimensional (discrete or continuous) action space and \textit{deterministic} environments. The extension to multidimensional action space problems and formulation for stochastic environments will be studied in our future work.


We propose two types of permissibility under SAP. Type 1 says that after an action $a_t$ is performed in a state $s_t$ and the agent reaches the new state $s_{t+1}$, the agent can decide whether $a_t$ is \textit{permissible} or \textit{not permissible} in state $s_t$. Type 2 says that even without performing $a_t$ in $s_t$, the agent can already decide whether $a_t$ is permissible or not in $s_t$. An action is \textit{not permissible} in a state if the action can never lead to an optimal accumulated reward over the long run and thus should not be tried. An action is \textit{permissible} if it is \textit{not known} to be non-permissible (i.e., a permissible action can still be non-permissible but it is not known). Clearly, avoiding a non-permissible action is not a greedy decision that counters RL's philosophy of sacrificing the immediate reward to optimize for the accumulated reward over the long run. Since type 2 permissibility is simple (we will see it in the experiments), we focus only on type 1. Type 1 is intuitive as we humans often encounter situations when we regret a past action, and based on it, we avoid making the same mistake in an identical or similar future situations. 

\begin{wrapfigure}{r}{0.19\textwidth}
    \vspace{-0.2cm}
	\centering
	\includegraphics[height=3 cm]{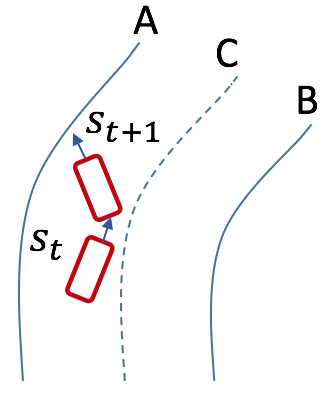}
	\caption{A lane keeping example in autonomous driving.}
	\label{task_example}
\end{wrapfigure}

The class of problems with the SAP property is quite large as most robot functions involving navigation or physical movements and even many games have this property (see Sec. 6). We use an example in autonomous driving (Figure 1) to illustrate type 1 permissibility. Here the car needs to learn appropriate steering control actions to keep it driving within a lane, called \textit{lane keeping}. A and B are the lane separation lines (track edges), and C is the center line (track axis) of the lane. We use the term ``\textit{track}" and ``\textit{lane}" interchangeably. The ideal trajectory for the car is the center line. At a particular time step $t$ the car is in state $s_t$ (Fig. 1). It takes an action $a_t$, i.e., turning the steering wheel counterclockwise for a certain degree. This action leads the car to state $s_{t+1}$. Note, $s_{t+1}$ is worse than $s_t$. Clearly, $a_t$ is non-permissible in $s_t$ as it would never lead to an optimal accumulated reward in the long run and thus should not have been taken (see footnote 3 for more explanations). 
\textbf{However, knowing $a_t$ is non-permissible in $s_t$ only after $a_t$ has been taken is not very useful. It is more useful if the information is used to help predict permissible and non-permissible actions in a state so that a permissible action can be chosen in the first place.} This is our goal, which is analogous to human learning as we use our knowledge about the environment to smartly choose permissible actions rather than blindly try all possibilities. Note, for type 2 permissibility, prediction is not needed (see Sec. 4).

For type 1 permissibility, we propose to use previous states, their actions, and the permissibility information of the \textbf{executed} actions to build a binary predictive model to infer the permisibility of a chosen action in current state and use it in action space exploration during RL training (see Sec. 4). A major advantage of the predictive model is that it is trained concurrently with the RL model and requires no human labeling of training data, which are obtained automatically during RL training by defining a type-1 \textit{Action Permissibility} (AP1) function and exploiting the SAP property (see Sec. 4.1). Type 2 permissibility is directly specified by defining a type-2 AP (AP2) function. The AP functions (AP1 \& AP2) are task/domain dependent and represent \textit{the encoded human knowledge (guidance) for exploring the concerned task}, provided by the user (developer)\footnote{One might argue that it requires deep domain expertise to specify an AP function. This may be true for complex environments. However, in practice the developer often has a good understanding of the environment/problem before he chooses to apply RL. Also, it is often possible to come up with sufficiently good AP function just based on commonsense for many tasks (like games, robot navigation, etc.), as we will discuss in Sec. 6. 
Moreover, the formulation of AP functions for similar environments are very similar, which reduces the requirement of intense analysis (see Sec 6).} 

For type 1 permissibility, as the agent experiences more states and actions during RL training and gathers knowledge (labels) of action permissibility, the predictive model becomes more accurate (stabilizes after some time), and thus, provides more accurate guidance over time, making RL training more efficient. One may ask here: what happens if the predictive model predicts wrongly? - there are two cases. First, if a non-permissible action is predicted as permissible in a state, it just results in some waste of time. After the action is performed, the agent will detect that the action is non-permissible and it will be added to the training data for the predictive model to improve upon in the next iteration. Second, if a permissible action is predicted as non-permissible, this is a problem as in the worst case, RL may find no solution. We solve this in Sec.~4.  

It is well-known that delayed or sparse rewards present a key challenge to many RL problems, as they slow down learning through long, uninformed explorations~\cite{marom2018belief}. A \textbf{Virtual Stopping} strategy is proposed to exploit the SAP property to deal with the challenge. It works as follows: once an non-permissible action is detected, the system virtually stops the training episode (without stopping for real) with a negative reward so that the system can be trained without delay or waiting for the termination of the whole episode.

In summary,  this paper makes three main contributions. 
\textbf{(1)} It identifies a special SAP property in a class of RL problems that can be leveraged to cut down the exploration space to markedly speed up RL training. To our knowledge, the property has not been reported before. 
\textbf{(2)} It proposes a novel approach to using the SAP property, i.e., building a binary predictive model to predict whether an action in a state is permissible or not ahead of time. \textbf{(3)} It proposes a virtual stopping strategy to maximally exploit the SAP property.
Experimental results show that the proposed approach can result in a large speedup in RL training.

\section{Related Work}
Many attempts have been made to speedup RL, e.g., searching possible parameters for finding the fastest quadrupedal locomotion~\cite{kohl2004policy}, adaptively balancing the exploration-exploitation trade-off~\cite{wu2017adaptive}, overcoming the exploration problem by providing demonstrations~\cite{nair2017overcoming}, data-efficient learning~\cite{deisenroth2011pilco}, learning policies and termination conditions of options~\cite{bacon2017option}, reward shaping for model-based learning~\cite{asmuth2008potential}, detecting symmetry and state equivalence~\cite{mahajan2017symmetry,girgin2010improving,osband2013more,bai2017efficient}, and those in \cite{narendra2016fast,duan2016rl,lipton2018bbq}. Although these works contribute in RL speed up, their problem set up and approaches differ significantly from ours. Recent advances in meta-RL achieved large policy improvements at test time with minimal sample complexity \cite{duan2016rl,wang2016learning,kulkarni2016hierarchical,finn2017model}, but they have not addressed the issue of exploration.

More closely related works to ours are those about action elimination. \cite{even2003learning} eliminated non-optimal actions with high probabilities for multi-armed bandits and tabular MDPs by learning confidence intervals around the value function. \cite{zahavy2018learn} eliminated actions in some text games. Although it discussed about using domain knowledge to eliminate actions, it only discarded invalid actions in a state, which can be detected easily in the text games based on knowledge about text processing. The work did not discuss how to extend to other types of RL tasks or non-promising actions. Our work is general. We not only introduce the concepts of the general SAP property and AP function, but also discuss many diverse RL experimental settings where such AP functions can be easily formulated for both continuous and discrete action spaces. Other less related works include integrating task hierarchies with offline RL~\cite{schwab2017offline}, using smooth dynamics of physical systems to enable the RL agent to never violate constraints for safe exploration~\cite{dalal2018safe}, avoiding catastrophic forgetting of dangerous states~\cite{lipton2016combating}, and exploiting affordance extraction~\cite{fulda2017can}. Compared to these, our work presents a general concept SAP for encoding prior application domain knowledge to help the RL agent in fast policy learning. We reported a priliminary research on this topic with a focus on autonomous driving in~\cite{mazumder2018action}. 

\cite{rosman2012good} proposed to use transfer learning to learn action priors 
from a set of different optimal policies from many tasks in the same state space to bias exploration away from less useful actions. \cite{abel2015goal} leveraged action priors from an expert or learned from related problems. Our work learns the action permissibility from the \textit{same} problem and is not based on transfer learning.  We focus on constraining the exploration space by using a property of the underlying task. 

Reward shaping \cite{ng1999policy,gao2015potential,marom2018belief} modifies reward function to accelerate RL training. Our work focuses on directly cutting the exploration space. Since our work involves prediction, it is remotely related to model-based RL \cite{deisenroth2011pilco,kamalapurkar2016model,berkenkamp2017safe,nagabandi2018neural,clavera2018model}. However, our work does not learn transition probabilities and is still model-free. 

\section{Background}
We incorporate the SAP guidance into two deep RL algorithms, Double Deep Q-Network (DDQN) \cite{van2016deep} and Deep Deterministic Policy Gradient (DDPG) \cite{lillicrap2015continuous}, which are based on Q-learning~\cite{watkins1992q}. Q-learning employs the greedy policy $\mu(s)=\arg\max_{a} Q(s,a)$. For continuous state space, it is performed with function approximators parameterized by $\theta^Q$, by minimizing the loss- $L(\theta^Q)=\mathbb{E}_{s_t\sim \rho^\beta, a_t\sim \beta, r_i\sim \mathcal{E}}$ [($Q(s_t, a_t) - y_t)^2$], where, $y_t=r(s_t, a_t) ~+$ \\ $\gamma Q(s_{t+1},\mu(a_{t+1}))$ and $\rho^\beta$ is the discounted state transition distribution for policy $\beta$. The dependency of $y_t$ on $\theta^Q$ is typically ignored.
\cite{mnih2015human,mnih2013playing} adapted Q-learning by using deep neural networks as non-linear function approximators and a replay buffer to stabilize learning, known as Deep Q-learning or DQN. \cite{van2016deep} introduced Double Deep Q-Network (DDQN) by introducing a separate target network for calculating $y_t$ to deal with DQN's over-estimation problem.

For continuous action space problems, Q-learning is usually solved using an Actor-Critic method, e.g., DDPG \cite{lillicrap2015continuous}. DDPG maintains an Actor $\mu(s)$ with parameters $\theta^\mu$, a Critic $Q(s, a)$ with parameters $\theta^Q$, and also a replay buffer $\mathcal{R}$ as a set of experience tuples ($s_t, a_t, r_t, s_{t+1}$) to store transition history for training. Training rollouts are collected with extra noise for exploration: $a_t = \mu(s)+\mathcal{N}_t$, where $\mathcal{N}_t$ is a noise process. In each training step, DDPG samples a minibatch of $N$ tuples from $\mathcal{R}$ to update the Actor and Critic networks and minimizes the following loss to update the Critic:
\vspace{-0.5mm}
\begin{equation} \label{eqn_1}
L(\theta^Q)=\frac{1}{N} \sum_i ~[y_i - Q(s_i, a_i)^2]
\vspace{-0.5mm}
\end{equation}
\noindent
where, $y_i=r_i + \gamma Q(s_{i+1},\mu(s_{i+1}))$. The Actor parameters $\theta^\mu$ are updated using the sampled policy gradient: 

$\nabla_{\theta^\mu} J=\frac{1}{N} \sum_i \nabla_a~[Q(s, a)|_{s=s_i,a=\mu(s)}~\nabla_{\theta^\mu} \mu(s)|_{s=s_i}]$.

\section{Proposed Technique}
The proposed framework consists of the SAP property, an action permissibility prediction model, and the integration of the predictive model in RL for guidance. 

\subsection{State-Action Permissibility (SAP)} 
Let $r:(\mathcal{S}, \mathcal{A}) \rightarrow \mathbb{R}$ be the reward function for a given MDP with state space $\mathcal{S}$ and action space $\mathcal{A}$. As mentioned before, we assume that the action space is one-dimensional (expressed by one variable) and the MDP is deterministic. 

\textbf{Definition 1} (\textit{permissible} and \textit{non-permissible action}): If an action $a_t$ in a state $s_t$ cannot lead to an optimal accumulated reward in the long run, the action is said to be \textit{non-permissible} in the state. If $a_t$ in $s_t$ is not known to be non-permissible, it is \textit{permissible}.

\textbf{Definition 2} (\textit{type 1 permissibility}): Let a state transition in a RL problem be ($s_t$, $a_t$, $r_t$, $s_{t+1}$). We say that the RL problem has the type 1 SAP property if there is a \textit{type 1 action permissibility} (AP1) function $f_1:(\mathcal{S}, \mathcal{A}, \mathcal{S}) \rightarrow \{0, 1\}$ that can determine whether action $a_t$ in state $s_t$ is \textit{permissible} [$f_1(s_t,a_t|s_{t+1})=1$] or \textit{non-permissible} [$f_1(s_t,a_t|s_{t+1})=0$] in $s_t$ after the action $a_t$ has been performed and the agent has reached state $s_{t+1}$.

\textbf{Definition 3} (\textit{type 2 permissibility}): Let the RL agent be in state $s_t$ at a time step $t$. 
We say that a RL problem has the type 2 SAP property if there is a \textit{type 2 action permissibility} (AP2) function $f_2:(\mathcal{S}, \mathcal{A}) \rightarrow \{0, 1\}$ that can determine whether an action $a_t$ in $s_t$ is \textit{permissible} [$f_2(s_t,a_t)=1$] or \textit{non-permissible} [$f_2(s_t,a_t)=0$] without performing $a_t$. 

Clearly, a permissible action may still be non-permissible, but it is not known. Both AP1 and AP2 functions may not be unique for a problem. Since type 2 permissibility (AP2) is simple, we focus only on type 1 permissibility (AP1). We illustrate AP1 using an example in the lane keeping task of self-driving cars.

\begin{wrapfigure}{r}{0.22\textwidth}
	\centering
	\vspace{-0.4cm}
	\includegraphics[height=2.9cm]{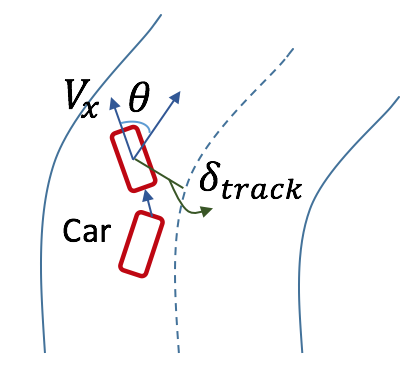}
	\vspace{-0.4cm}
	\caption{\small Parameters of the reward function in Equation 2.}
	\vspace{-0.4cm}
	\label{task_example}
\end{wrapfigure}

\textbf{Example 1.}  Let, at any given time \textit{in motion}, $\theta$ be the angle between the car's direction and direction of the track (lane) center axis, $V_x$ be the car speed along the longitudinal axis and $\delta_{track}$ be the distance of the car from the track center (center line) (see Figure 2). 
Given this setting, we use the following reward function (an improved version\footnote{https://yanpanlau.github.io/2016/10/11/Torcs-Keras.html} of that in \cite{lillicrap2015continuous}) for the lane keeping task: 
\vspace{-0.5mm}
\begin{equation} \label{eqn_r2}
r=V_x (\cos\theta - |\sin\theta| - \delta_{track})
\vspace{-0.5mm}
\end{equation}

The car gets the maximum reward ($V_x$) only when it is aligned with the track axis and $\delta_{track}=0$. 

Let $\delta_{track, t}$ and $\delta_{track, t+1}$ be the distance of the car from the lane center line (track axis) corresponding to state $s_t$ and state $s_{t+1}$ respectively. The following is an AP1 function: 

\vspace{-2mm}
\begin{equation}\label{key}
f_1(a_t, s_t | s_{t+1}) = \begin{cases}
0 & \text{if $\delta_{track, t+1} - \delta_{track, t} > 0$}\\
1 & \text{Otherwise}
\end{cases}
\vspace{-0.5mm}
\end{equation}

This AP1 function says that any action results in the car to move further away from the track axis (center line) is not permissible.\footnote{In a sharp turn, if the curvature of the lane is too large, human drivers may drive to the outer side of the lane or even cut into the neighboring lane before the turn. This is usually because we drive too fast. To deal with it, we need speed control. 
In some cases, we have to go out of lane/track to avoid a collision or to make a U-turn. In such a case, the self-driving system usually dynamically generates a virtual lane for the car to travel. With the virtual lane, our method still applies. We leave these and speed control (which adds another action dimension) to our future work.}
This clearly satisfies the type 1 SAP property because without performing the action, one will not know whether the action is permissible or not. 

\subsection{Learning Type 1 AP (AP1) Predictor} 
AP1 function \textbf{only gives knowledge} about the permissibility of ``\textbf{\textit{executed}}" actions, which serves as labels for the actions. 
\textbf{Our goal} is to \textit{use these past action experiences labeled by the AP1 function to learn an AP1 predictor so that the agent can look ahead the permissibilty of a possible action in the current state}.

As indicated earlier, AP1 prediction is a binary classification problem. Given the current state $s$ and an action $a$, AP1 predictor predicts whether $a$ is permissible or not permissible in $s$. \textit{Note that AP1 predictor is a learned predictive model, which is different from the user-provided AP1 function.} The labeled training data for building AP1 predictor is produced by AP1 function $f_1$. 
Each training example consists of a state and an action \textit{taken} in the state with its class (permissible or non-permissible). After some initial steps of RL, a set of training examples for training AP1 predictor is collected. 

Since AP1 predictor training is done continuously along with the RL training, to manage the process and the stream of new training examples, we maintain a training buffer $\mathcal{K}$ like the replay buffer $\mathcal{R}$ in \cite{lillicrap2015continuous,mnih2013playing} to train the AP1 predictor. Given a RL experience tuple ($s_t, a_t, r_t, s_{t+1}$) at time step $t$, we extract the tuple ($s_t, a_t, l(a_t)$) and store it in $\mathcal{K}$. Here, $l(a_t)$ is the class label for $a_t$ in $s_t$, permissible (+ve class) or non-permissible (-ve class), and is inferred using the AP1 function $f_1$. $\mathcal{K}$ is finite in size and when it gets full, newer tuples replace oldest ones having the same class label $l(a_t)$.

We train AP1 predictor $E$ with a balanced dataset at a time step $t$ as follows. For step $t$, if both the number of +ve and -ve tuples (or examples) in $\mathcal{K}$ are at least $N_E/2$ (ensures $N_E/2$ +ve and $N_E/2$ -ve examples can be sampled from $\mathcal{K}$), we sample a balanced dataset $\mathcal{D}_E$ of size $N_E$ from $\mathcal{K}$. Then, we train the neural network AP1 predictor $E$ with parameter $\theta^{E}$ using $\mathcal{D}_E$. Note that, AP1 predictor is just a supervised learning model. We discuss the network architecture of AP1 predictor used for our experiments in Supplementary. For training $E$, we minimize L2-regularized (with regularization parameter $\lambda$) cross-entropy loss:
\vspace{-0.1cm}
\begin{equation} \label{eqn_1}
\small
\begin{split}
L(\theta^E)= -\frac{1}{N_E} \sum_{(s_i, a_i, l(a_i)) \in \mathcal{D}_E} [~l(a_i)~log~E(s_i, a_i)+ \\
(1-l(a_i))~log~(1-E(s_i, a_i))] + \frac{\lambda}{2} \sum \norm{\theta^E}^2_2
\end{split}
\end{equation}

\setlength{\textfloatsep}{0.1cm}
\setlength{\floatsep}{0.1cm}
\begin{algorithm}[t!]
	\small
	\caption{\small AP Guided Action Selection}
	\label{alg:example}
	\begin{flushleft}
		\textbf{Input:} Current state $s_t$; $\mu(s)$ as RL Action selection model (e.g, Actor network in DDPG or DDQN); AP1 predictor $E(s, a)$; current time step $t$; Observation time step threshold $t_o$; Exploration time step threshold $t_e$ ($> t_o$); probability threshold $\alpha_e$ and $\alpha_{tr}$ ($> \alpha_e$) for consulting $E$ and $v_{acc}^{t-1}(E)$ as validation accuracy (computed on hold out examples/tuples in $\mathcal{K}$) of $E$ computed at time step $t-1$. \\
		\textbf{Output:} $a_t$: action selected for execution in $s_t$
	\end{flushleft}
	\vspace{0.1cm}
	\begin{algorithmic}[1]
		\State Select action $a_t=\mu(s_t)$ for $s_t$   
		\If{$t \leq t_e$}  \Comment{Exploration phase}
		\State $a_t$ = Exploration($a_t$)  \Comment{Use Noise process for DDPG and exploration strategy for DDQN} 
		\EndIf
		\If{$t > t_o$}    \Comment{Start AP1 guidance when $t > t_o$ (observation phase is over)}
		\If{$t > t_e$ and $v_{acc}^{t-1}(E) \geq \delta_{acc}$}
		\State Set $\alpha = \alpha_{tr}$ 
		\Else 
		\State Set $\alpha = \alpha_e$ 
		\EndIf
		\State  $\hat{l(a_t)} \leftarrow E(s_t, a_t|\theta^E)$ 
		\If{$\hat{l(a_t)}$ is -ve and $Uniform(0, 1) < \alpha$}
		\State Select Candidate Action Space $\mathcal{A}_{s_t}$ from $\mathcal{A}$ and build $D_{s_t}$ as \{($s_t, a)~|~a \in \mathcal{A}_{s_t}$\} \Comment{For DDPG, we sample $\mathcal{A}_{s_t}$ using low-variance uniform sampling from $\mathcal{A}$ and for DDQN, $\mathcal{A}$ being finite, $\mathcal{A}_{s_t}$ = $\mathcal{A}-\{a_t\}$}
		\State $\mathcal{A}_P(s_t) =\{a ~|~ E(s_t, a)$ is +ve, $(s_t, a) \in  D_{s_t}\}$  
		\If{$\mathcal{A}_P(s_t) \neq \emptyset$} 
		\State Randomly sample $a_t$ from $\mathcal{A}_P(s_t)$  \Comment{$a_t$ is sampled from predicted permissible action space}
		\EndIf
		\EndIf
		\EndIf
		\State Return $a_t$
	\end{algorithmic}
	\normalsize
\end{algorithm}
\setlength{\textfloatsep}{0.4cm}
\setlength{\floatsep}{0.4cm}

\vspace{-0.1cm}
\noindent We discuss the use of AP1 predictor in a RL model training below.

\subsection{Guiding RL Model with AP1 Predictor}
Our AP1 predictor can work with various RL models. In this work, we incorporate it into two deep RL models: DDPG~\cite{lillicrap2015continuous} and DDQN~\cite{van2016deep} (see Sec. 3). We chose DDPG (DDQN) as it is a state-of-the-art algorithm for solving continuous state and continuous (discrete) action space RL problems. Our integrated algorithm of DDPG and AP1 predictor $E$ is called DDPG-AP1, and of DDQN and $E$ is called DDQN-AP1. Training of Actor $\mu$ and Critic $Q$ of DDPG-AP1 (and that of Q-network for DDQN-AP1) is identical to the DDPG (DDQN) algorithm (see Sec. 3) \textbf{except} one modification (see Sec. 4.4). 
We discuss how a trained $E$ (say, up to time step $t$) helps in guided action exploration onwards.

Algorithm 1 presents the action selection process of DDPG-AP1/DDQN-AP1. Given the trained AP1 predictor $E$ at time step $t$, action selection for $s_t$ works as follows: Initially, DDPG-AP1/DDQN-AP1 selects action $a_t$ randomly from action space $\mathcal{A}$ via an exploration process upto $t \leq t_e$ steps (state transitions). This phase is usually called the \textit{Exploration Phase} in RL literature (lines 2-4). After $t > t_e$, the exploration process is not used further. For $t > t_o$, AP1 guided action selection process (lines 6-18) is enabled, where $t_o < t_e$.  Initially, when the RL agent starts learning, the tuples stored in $\mathcal{K}$ are few in number, not enough to build a good AP1 predictor. Moreover, $E$ needs a diverse set of training examples/tuples to learn well. Thus, for the initial set of steps ($t \leq t_o$), AP1 guidance is not used, but $E$ is trained. This phase is called \textit{Observation Phase}, which is the initial $t_o$ steps of the exploration phase. 

After the observation phase is over ($t > t_o$), for $t_o < t \leq t_e$, AP1 Predictor based guidance (Lines 5-19) and exploration process (Lines 2-4) work together, and for $t > t_e$, only AP1 based guidance (Lines 5-19) works. We call this phase the \textit{Learning/Training Phase}, where no more action exploration is done using the exploration process (lines 2-4).
For $t > t_o$, AP1 based guidance (lines 5-19) works as follows: the action selected in lines 1-4 is fed to $E$ for AP1 prediction with probability $\alpha$. $\alpha \in [0, 1)$ controls the degree by which DDPG-AP1/DDQN-AP1 consults $E$. As mentioned in Sec. 1, since AP1 predictor is not 100\% accurate, we need to deal with the case where a permissible action is predicted as non-permissible (false negative) (false positive is not an issue) as in the worst case, RL may not find a solution. We deal with the problem by letting the Actor to listen to $E$ for $\alpha$\% of the time. This probability ensures that the RL model executes the action generated by itself (including some false negatives) on environment $(1-\alpha)$\% of the time. Moreover, setting appropriate $\alpha$ also allows the RL model to experience some bad (non-permissible actions) experiences through out its training, which stabilizes RL learning. 
For $t_o < t \leq t_e$, we set $\alpha$ to a small value $\alpha_e$ to encourage more random exploration. Once exploration phase is over ($t > t_e$), $\alpha$ set to a bigger value $\alpha_{tr}$ ($> \alpha_e$) so that the agent often consults $E$.

In line 11, if $a_t$ is predicted as permissible by $E$, we skip lines 12-19 and action $a_t$ (selected in lines 1-4) gets returned (and executed on the environment). Otherwise, in line 13, the RL model selects a candidate action set $\mathcal{A}_{s_t}$ for $s_t$ of size $N$ from the action space $\mathcal{A}$ and finds a permissible action for the current state $s_t$. For DDPG-AP1, the action space being continuous, we estimate permissible action space by sampling an action set $\mathcal{A}_{s_t}$ for $s_t$ of size $N$ from $\mathcal{A}$ using low-variance uniform sampling. That is, $\mathcal{A}$ is first split into $N$ equal sized intervals and an action is sampled from each interval uniformly to produce a set of sampled actions $\mathcal{A}_{s_t}$ for state $s_t$. This ensures that actions are sampled uniformly over $\mathcal{A}$ with variances between consecutive samples being low. Thus, any action in $\mathcal{A}$ will be equally likely to be selected if it is predicted to be permissible (+ve) by $E$ (line 14). The uniform sampling is used because AP1 predictor does not learn the value function. It is logical to estimate the action permissibility space and let RL find the best policy from the permissible space, rather than choosing the best action greedily based on AP1 prediction score.

\begin{figure*}
    \centering
	\stackunder[5pt]{\includegraphics[height=2.57cm]{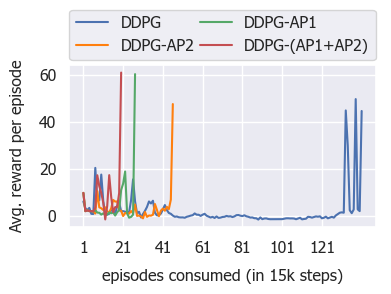}}{}%
	\stackunder[5pt]{\includegraphics[height=2.57cm]{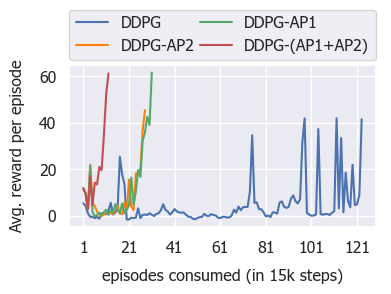}}{}%
	\stackunder[5pt]{\includegraphics[height=2.57cm]{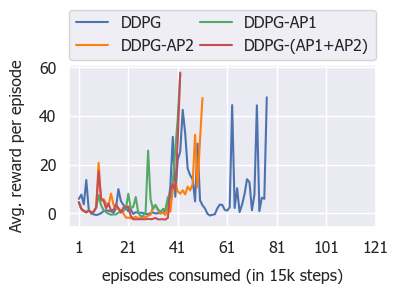}}{} 
	\stackunder[5pt]{\includegraphics[height=2.57cm]{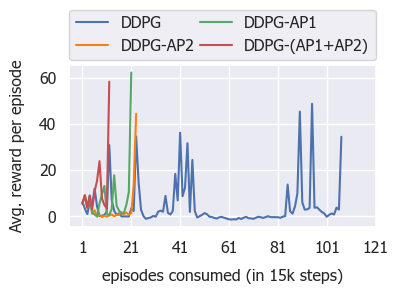}}{}%
	\stackunder[5pt]{\includegraphics[height=2.57cm]{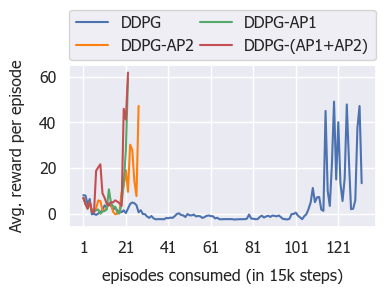}}{}%
	\label{task_example}
	\vspace{-0.15cm}
	\caption{\small Avg. reward per episode for DDPG and DDPG-AP variants on the lane keeping task over five training experiments with different random seeds. 1.2k exploration steps were used in training.}
	\vspace{-0.25cm}
\end{figure*}

For DDQN-AP1, since the action space is finite, $\mathcal{A}_{s_t}= \mathcal{A}-\{a_t\}$. Once $\mathcal{A}_{s_t}$ is selected, RL forms a dataset $D_{s_t}$ by pairing $s_t$ with each $a \in \mathcal{A}_{s_t}$ and feeds $D_{s_t}$ to $E$ in a single batch to estimate a permissible action space for $s_t$ as $\mathcal{A}_P(s_t)$ (line 16). Here +ve means permissible. Next, RL model randomly samples an action $a_t$ from $\mathcal{A}_P(s_t)$ (line 16) and executes it on environment (not in Algorithm 1). If $\mathcal{A}_P(s_t)=\emptyset$, the original $a_t$ (selected in lines 1-4) is returned and gets executed. $\alpha_e$, $\alpha_{tr}$ and $\delta_{acc}$ are chosen empirically (discussed in Appendix D). 

\subsection{Virtual Stopping} 
It is well-known that delayed and sparse rewards present a key challenge in many RL problems, as they slow down learning through long, uninformed explorations. In the proposed framework, when a non-permissible action is detected, a negative reward can be given as the action will not lead to an optimal reward in the long run. However, in a delayed reward RL problem, the RL model doesn't learn the knowledge \textbf{immediately}. Rather, it slowly learns through credit assignment in a delayed learning process over a long horizon (e.g., after a game/episode ends), which diminishes the advantage of the AP-based guidance. 

To alleviate this problem, we propose \textbf{virtual stopping} (VS) to enable credit assignment over a shorter horizon to boost RL training. VS works as follows: whenever the agent executes a non-permissible action $a_t$ and receives an non-permissible experience, it assumes that it has (virtually) ended the episode with a negative reward (as the solution in the current episode can no longer lead to an optimal reward) and "end of episode" flag is set true with the experience stored in the replay buffer. Note that the episode does not end in real, and starting from the current state a new virtual episode starts. VS just enables training over shorter horizons. This drastically accelerates RL training.

Note that, for delayed and sparse reward RL problems, without VS, AP-based guidance will not be very effective. This is because AP-based guidance gets the agent to mostly perform permissible actions compared to non-permissible ones during training, but does not learn the consequence of taking non-permissible actions immediately in a given state, which can cause biased training.

\section{Experiments}
This section evaluates the proposed DDPG-AP and DDQN-AP techniques
on three RL tasks and analyze their learning performances, and compare them with the baseline DDPG / DDQN. We discuss the details of the task environments, RL model architectures and hyper-parameter settings in Appendix.

\subsection{Lane Keeping Task}


We use an open-source, standard autonomous driving simulator TORCS \cite{loiacono2013simulated} following \cite{sallab2016end,sallab2017deep} for both learning and evaluation. Figure 3 shows a snapshot of the TORCS simulator. We used five sensor readings to represent the state vector which we found are sufficient for learning good policies in diverse driving situations. The goal of our experiment is to assess how well the driving agent has learned to drive to position itself on the track/lane axis. By no means are we trying to solve the whole self-driving problem, which is much more complex (see footnote 3). Our model and baselines thus focus on predicting the right steering angle for lane keeping while driving with a default speed. \textit{During training, whenever the car goes out of the track, we terminate the current episode and initiate a new one.} We use five diverse road tracks in our experiments (See Section A in Appendix for more details).

\vspace{0.05cm}
\textbf{Compared Algorithms.} We compare our DDPG-AP (AP1 and AP2) models with the baseline \textbf{DDPG} algorithm (without any action selection guidance). Here, we also propose a type-2 AP guidance based on some driving characteristics. 

\textbf{DDPG-AP1.} An extension of DDPG that uses type 1 AP function (as proposed in Eqn. 3) and AP1 predictor for guidance.

\textbf{DDPG-AP2.} An extension of DDPG that applies the following two AP2 functions (or constraints): (1) If the car is on the left of track center and current action $a_t > a_{t-1}$ (previous action), instead of applying $a_t$, it samples actions uniformly from (-1.0, $a_{t-1}$)\footnote{Steering values -1 and +1 mean full right and left respectively.} and thus, avoids taking any left turn further. Similarly, (2) if the car is on the right of track center and $a_t < a_{t-1}$, then sample actions from ($a_{t-1}$, 1.0) and avoid turning right further. Otherwise, the car executes $a_t$. These constraints are applied only when $\delta_{track,t}-\delta_{track,t-1}>0$, i.e., when the car moves away from the track center due to $a_{t-1}$. If the car is moving closer to the track center, it is permissible. This model does not use AP1 prediction.

\textbf{DDPG-(AP1+AP2).} Version of DDPG where we combine our DDPG-AP1 (type 1 AP) and DDPG-AP2 (type 2 AP). Here, we learn AP1 predictor for training DDPG-(AP1+AP2) due to the use of type 1 permissibilty.

In the lane keeping task, as the reward is immediate, virtual stopping is not needed for training DDPG-AP1, DDPG-AP2 or DDPG-(AP1+AP2).

\textbf{Results and Analysis.} Fig. 3 shows the average number of episodes consumed by each algorithm in 15k training steps (or state transitions). \textbf{If an algorithm consumes less number of episodes, it means the algorithm learns quicker to keep the car moving without going out of track.} From Fig. 3, we observe that DDPG took more than 100 episodes (on avg.) to learn to drive for a considerable amount of distance. However, the sharp falls in the average reward and initiation of a new episode indicate that the learning is yet not stable. Among all, DDPG-AP1 and DDPG-(AP1+AP2) learn very quickly, and \textbf{as their curves do not fall down, which indicates the car has never gone out of track after 20$^{th}$ [for DDPG-(AP1+AP2)] and 26$^{th}$ (for DDPG-AP1) episodes on average, which also indicates the said algorithms have already learned a stable policy for driving}.

We also evaluated AP1 predictor's validation accuracy ($v_{acc}^{t-1}(E)$, see Algorithm 1) and found that the accuracy always stays above 70\% during initial training steps and stabilizes with an average of 80\% as shown in Fig. 4. As we can see here, the  validation accuracy of the AP classifiers increases over time with the incoming examples labeled by the corresponding AP function and then, gradually saturates to fairly high accuracy, denoting the learning has become stable. This suggests that, we do not need to train the predictor for all steps during the whole training period. During RL training, we postpone the training of the predictor, whenever the validation accuracy is above a threshold $\delta_{acc}$ [see Line 6, Algorithm 1] and resume its training whenever the validation accuracy falls below the threshold utill it again goes above $\delta_{acc}$ after an initial phase of training (until the validation accuracy starts to stabilize). Here, $\delta_{acc}$ is considered as the threshold for predictor's reliability.

\begin{wrapfigure}{r}{0.22\textwidth}
    \centering
    \vspace{-0.2cm}
	\includegraphics[height=3cm]{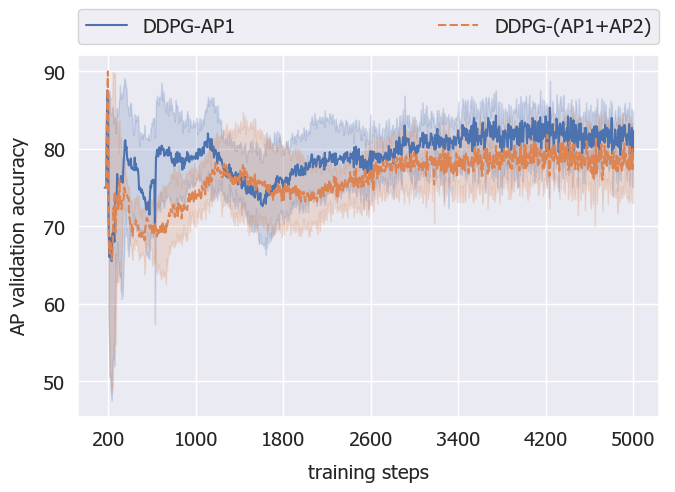}
	\label{task_example}
	\vspace{-0.4cm}
	\caption{Validation accuracy of AP1 predictor/classifier over training steps for lane keeping task.}
	\vspace{-0.2cm}
\end{wrapfigure}

Table 1 shows the performance of the algorithms on unseen test tracks for 3k and 15k training steps. We use each algorithm to drive the car for one lap of each track and report the total reward obtained by the algorithm with the lap completion information [(lap?) denotes whether the car has completed the lap or not]. Considering the results for the 3k training steps (which is very few for learning a stable policy), we see that DDPG and DDPG-AP2 have not learned to make the car complete one lap for all test tracks. Both DDPG-AP1 and DDPG-(AP1+AP2) perform much better in term of lap completion. For 15k training steps, we see that all algorithms except DDPG  have learned to keep the car on track for all test tracks. The highest total reward values and lap? information in DDPG-(AP1+AP2) (considering all test tracks) indicate that DDPG-(AP1+AP2) has learned to find the most general policy quickly compared to the others. Although DDPG-AP2 was competitive with DDPG-AP1 in 3k training steps, the rewards obtained in 15k are less than those for DDPG-AP1 and DDPG-(AP1+AP2). This shows that the policy learned by DDPG-AP2 is sub-optimal.

\begin{table}[t!]
	\scriptsize
	\centering
	\caption{Performance of DDPG and DDPG-AP variants on 4 different test tracks.}
	\vspace{-0.1cm}
	\label{my-label}
	\begin{tabular}{|p{1.4cm}|c|c|c|c|}
		\hline
		& \multicolumn{1}{c|}{DDPG}                                                                                                                                                            & \multicolumn{1}{c|}{DDPG-AP2}                                                                                  & \multicolumn{1}{c|}{DDPG-AP1}                                                                                   & \multicolumn{1}{c|}{DDPG-(AP1+AP2)}                                                                            \\ \hline
		\multicolumn{1}{|c|}{Test track}                           &  \begin{tabular}[c]{@{}c@{}}Reward~ (Lap?)\end{tabular} &   \begin{tabular}[c]{@{}c@{}}Reward~ (Lap?)\end{tabular} &  \begin{tabular}[c]{@{}c@{}}Reward~ (Lap?)\end{tabular} &  \begin{tabular}[c]{@{}c@{}}Reward~ (Lap?)\end{tabular} \\ \hline
		\multicolumn{5}{|c|}{Training track:  Wheel-2   {[}After 3k steps training{]}}                                                                                                                                                                                                                                                                                                                                                                                                                                                                                                                                                                \\ \hline
		E-road        & 4460.96 ~(N)    & 53371.60 ~(Y)  & 53189.98 ~(Y)   & \textbf{54667.40} ~(Y)  \\ 
		Spring        & 8258.29 ~(N)    & 162875.81 ~(N)  & 368724.99 ~(Y)  & \textbf{371795.67} ~(Y) \\ 
		CG Track 3    & 1574.31 ~(N)    & 46948.97 ~(Y)  & 46364.52 ~(Y)  &  \textbf{48199.51} ~(Y) \\ 
		\begin{tabular}[c]{@{}l@{}}Oleth Ross \end{tabular} & 8784.68 ~(N) & 105419.94 ~(Y)                        & 103557.34 ~(Y) & \textbf{107168.99} ~(Y)                                              \\ \hline
		\hline
        \multicolumn{5}{|c|}{Training track:  Wheel-2    {[}After 15k training steps{]}} \\ \hline
		 E-road   & 40785.05 ~(Y)   & 53653.04 ~(Y)  & 56217.72 ~(Y)  & \textbf{56333.61} (Y) \\ 
		 Spring   & 117519.69 ~(N)  & 368559.39 ~(Y) & 382746.63 ~(Y) & \textbf{383541.04} ~(Y)   \\ 
		 CG Track 3  & 37011.54 ~(Y) & 46975.31 ~(Y) & 49085.96 ~(Y) & \textbf{49535.38} ~(Y)  \\ 
		 Oleth Ross  & 86275.80 ~(Y) & 105584.83 ~(Y) & 109506.41 ~(Y) & \textbf{110384.84} ~(Y) \\ \hline
	\end{tabular}
	\vspace{-0.1cm}
\end{table}

\subsection{Flappy Bird} 

Since this is a discrete action space problem, we use the RL network DDQN (see Sec. 3). 
We use the open source pygame version of Flappy Bird\footnote{github.com/yenchenlin/DeepLearningFlappyBird} for evaluation. The goal here is to make a bird learn to fly and navigate through gaps between pipes (see Fig. 5), where the allowed actions are \{\textit{flap}, \textit{no flap}\}. The \textit{flap} action causes an increase in upward acceleration and \textit{not flap} makes the bird fall downward due to gravity. 

\textbf{AP Function.} According to the Flappy bird game setting, whenever the bird flaps, it accelerates upward by 9 pixels and if it does not flap it accelerates downward by 1 pixel. These are the two possible discrete actions. An optimal solution for the game is when the expected trajectory of the bird follows the midway of the pipe gap. To achieve this and make each move safer (less prone to crashing the pipe), the bird should accelerate downward by some steps before the next flap. Thus, if the bird is above the next pipe gap center line, a flap increases the chance of hitting the upper pipe compared to that when below the gap center line. Also, if the bird is below the top surface of next lower pipe, not flapping causes the bird to fall down and reduces the possibility of reaching to the next pipe gap on the next flap without hitting the lower pipe. Based on this observation, a type-2 AP function can be formulated as follows:

\begin{wrapfigure}{r}{0.17\textwidth}
    \centering
    \includegraphics[height=4cm]{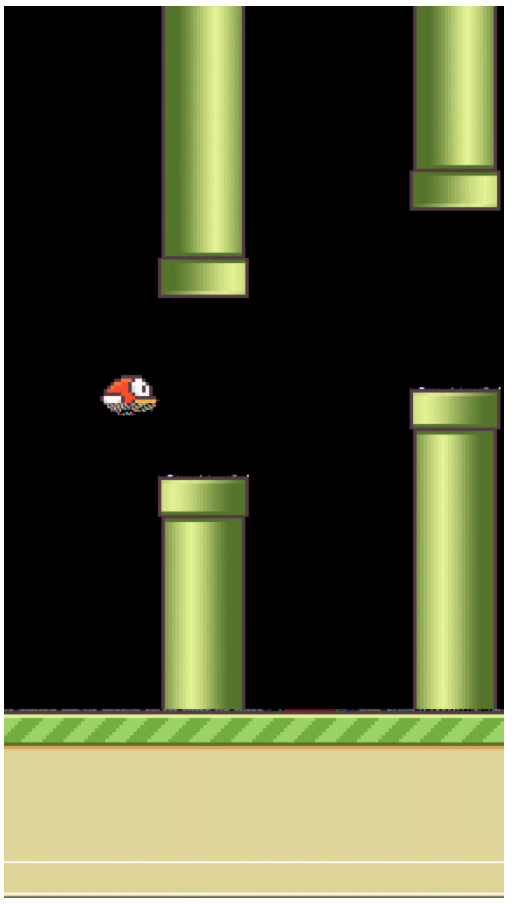}%
	\label{task_example}
	\vspace{-0.1cm}
	\caption{\small Snapshot of Flappy Bird game.}
	\vspace{-0.25cm}
\end{wrapfigure}

Let $c$ be the horizontal line that passes through the mid point of the next pipe gap and $\delta_c^t$ be the vertical distance of the agent (bird) from $c$ at state $s_t$. If $\delta_c^t > 0$, the bird lies above the gap center line $c$ and vice versa. Also, let $l$ be the horizontal line that passes through the next lower pipe's Y coordinate (i.e., Y-coordinate of next gap's bottom left point) and $\delta_l^t$ be the vertical distance of the agent (bird) from $l$ at state $s_t$. If $\delta_l^t > 0$, the bird lies above the lower pipe top surface line $l$ and vice versa. Then a type-2 AP (AP2) function can be defined as:
\vspace{-0.2cm}
\begin{equation}\label{key}
\small
f_2(a_t, s_t) = \begin{cases}
0 & \text{if $C_1$ or $C_2$}\\
1 & \text{Otherwise}
\end{cases}
\end{equation}

\begin{figure*}
    \centering
	\captionsetup{justification=centering}
	\stackunder[5pt]{\includegraphics[height=2.57cm]{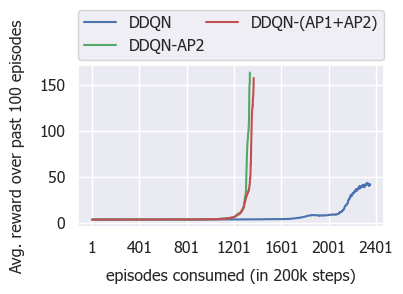}}{}%
	\stackunder[5pt]{\includegraphics[height=2.57cm]{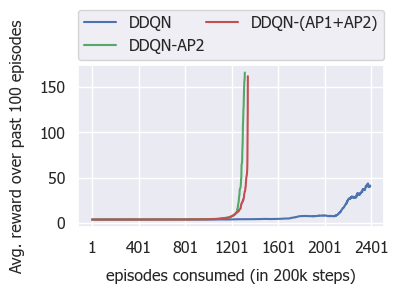}}{}%
	\stackunder[5pt]{\includegraphics[height=2.57cm]{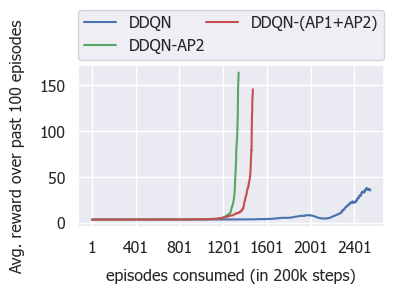}}{} 
	\stackunder[5pt]{\includegraphics[height=2.57cm]{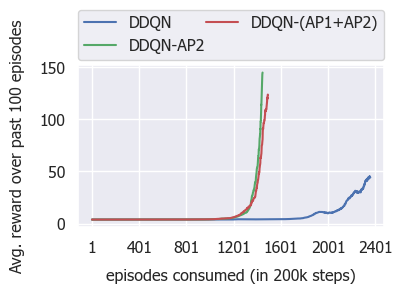}}{}%
	\stackunder[5pt]{\includegraphics[height=2.57cm]{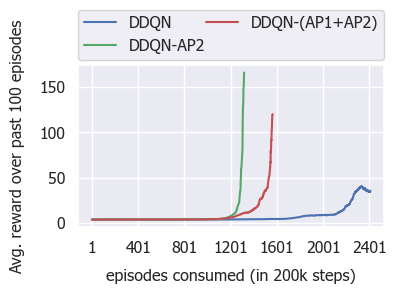}}{}%
	\vspace{-0.1cm}
	\label{task_example}
	\vspace{-0.15cm}
	\caption{\small Avg. reward over past 100 episodes for DDQN and DDQN-AP variants on Flappy bird over five training experiments (60k exploration steps) with different random seeds.}
	\vspace{-0.2cm}
\end{figure*}

\noindent 
where $C_1$ = \{$\delta_c^t > 0$, $a_t = ``flap"$\} and $C_2$ = \{$\delta_l^t < 0$, $a_t= ``no flap"$\}. $C_1$ says that when the bird is above $c$, performing action ``\textit{flap}" that increases vertical acceleration (causing the bird move further up) is non-permissible. Similarly, when the bird is below $l$, performing ``\textit{no flap}" results in the bird to move further down and so is non-permissible ($C_2$). Here whether an action $a_t$ satisfies any condition in \{$C_1$, $C_2$\} can be determined at $s_t$. Thus, $f_2$ in equation 1 indicates type 2 permissibility.

However, even if the bird is above $l$, repeated ``\textbf{no flap}" action can cause the bird to hit surface of lower pipe specially when the bird is within the pipe gap. And only when the bird crashes at $s_{t+1}$, we can conclude that ``\textit{no flap}'' in $s_t$ was non-permissible. Thus, we introduce a type 1 AP (AP1) function (see Section 4.1) as follows:
\vspace{-0.2cm}
\begin{equation}\label{key}
\small
f_1(a_t, s_t | s_{t+1}) = \begin{cases}
0 & \text{if $C_3$}\\
1 & \text{Otherwise}
\end{cases}
\end{equation}
where $C_3$=\{$\delta_c^{t+1} < 0$, $a_t=``no flap$", $s_{t+1}~=~crash$\}. $C_3$ indicates whether the bird has crashed to lower pipe top surface in state $s_{t+1}$ due to ``\textit{no flap}'' in $s_t$. Thus, a new and stronger AP function (AP1+AP2) can be designed that combines $f_1$ and $f_2$ involving $C_1$, $C_2$ and $C_3$ [Eqn. 7], which covers all our action non-permissibility cases for Flappy bird:
\begin{equation}\label{key}
\small
f(a_t, s_t) = \begin{cases}
0 & \text{if $C_1$ or $C_2$ or $C_3$}\\
1 & \text{Otherwise}
\end{cases}
\end{equation}

\textbf{Compared Algorithms.} We compare DDQN-AP (AP1 and AP2) models below with the original \textbf{DDQN} algorithm. 

\textbf{DDQN-AP1.} An extension of DDQN that uses only $C_3$ as AP function ($f_1$). Permissibility guidance for $f_1$ is very week as non-permissible experiences are only accumulated when the bird crashes the lower pipe top surface. As this version does not result in significant speedup, we exclude it from discussion.

\textbf{DDQN-AP2.} An extension of DDQN that applies the type 2 AP function $f_2$ involving \{$C_1$, $C_2$\} and is a much stronger DDQN-AP variant compared to DDQN-AP1.

\textbf{DDQN-(AP1+AP2).} Version of DDQN where we combine our DDQN-AP1 and DDQN-AP2. Here, we learn the AP predictor to guide the learning of RL model, and use combined AP function (in Eqn. 7) to label the permissibility of an action taken.

In Flappy Bird, as the agent receives a delayed reward for crossing or crashing the pipe, and the same reward (0.1) for all other actions for being alive, we use virtual stopping (VS) in DDQN-AP2 and DDQN-(AP1+AP2).

\textbf{Results.} We trained DDQN-AP2, DDQN-(AP1+AP2) and DDQN for 200k steps with $\epsilon$-greedy strategy and conducted two experiments. In the first experiment, we used 30k exploration time steps and in the second experiment, we used 60k exploration time steps. 1000 initial observation steps were used in both cases. Fig. 6 shows the average rewards over past 100 training episodes (or games) for DDQN and DDQN-AP variants and episodes consumed by each algorithm in 200k training steps (state transitions)\footnote{Note, for lane keeping and flappy bird, as an episode can be infinitely large, the avg. reward curve may not flatten. Flattening of avg. reward curve means the learned policy always makes the agent to end episode after a certain number of steps.}.

\begin{wrapfigure}{r}{0.22\textwidth}
    \centering
    \vspace{-0.35cm}
	\includegraphics[height=3cm]{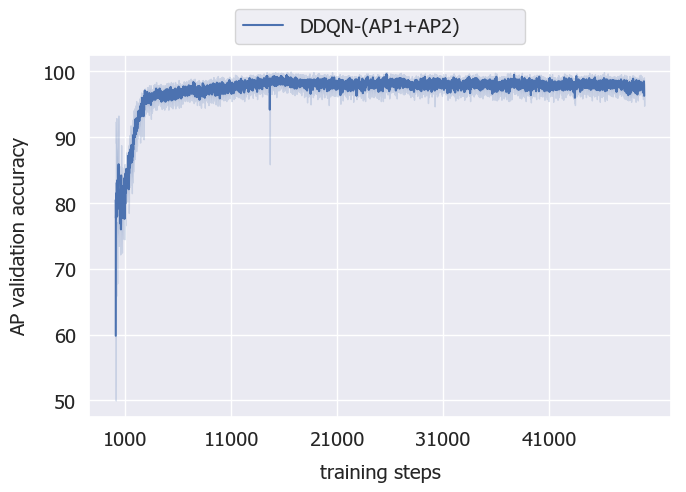}
	\label{task_example}
	\caption{Validation accuracy of AP1 predictor/classifier over training steps for flappy bird (for the 60k exploration training experiment).}
	\vspace{-0.2cm}
\end{wrapfigure}

From Fig. 6, we see that both DDQN-AP2 and DDQN-(AP1+AP2) train more rapidly than DDQN, which is reflected in the escalating growth of the avg. reward curve. DDQN-AP variants avoid collision with pipes for a longer period of time, consuming less episodes compared to DDQN. Although both DDQN-AP2 and DDQN-(AP1+AP2) learns rapidly, the learning of DDQN-(AP1+AP2) is more stable due to the incorporation of type-1 AP function (Eqn. 7), covering all non-permissibility cases, which is reflected in its test performance (See Table 2). We also noted that AP predictor's validation accuracy always stays above 90\% during training and stabilizes with 97.8\% on average as shown in Fig. 7.

\begin{table}[t!]
	\scriptsize
	\centering
	\caption{\small Average test scores over 50 games (episodes) of DDQN, DDQN-AP2 and DDQN-(AP1+AP2) on hard difficulty level (pipe gap = 100) of the Flappy bird game. The 50 games consists of five test experiments conducted with five different random seeds (10 games have been played in each test experiment by the model trained on a given random seed).}
	\begin{tabular}{|c|c|c|c|c|c|c|}
		\hline
		\multirow{2}{*}{\begin{tabular}[c]{@{}c@{}}training\\ steps\end{tabular}} & \multicolumn{3}{c|}{exploration steps = 30k}                                   & \multicolumn{3}{c|}{exploration steps = 60k}                                   \\ \cline{2-7} 
		& DDQN  & \begin{tabular}[c]{@{}c@{}}DDQN-\\AP2 \end{tabular} & \begin{tabular}[c]{@{}c@{}}DDQN-\\ (AP1+AP2)\end{tabular} & DDQN  & \begin{tabular}[c]{@{}c@{}}DDQN-\\AP2\end{tabular} & \begin{tabular}[c]{@{}c@{}}DDQN-\\ (AP1+AP2)\end{tabular} \\ \hline
		100k                                                                      & 20.96   & 55.22     & 165.4                                                     & 1.68   & 49.98     & 44.48                                                      \\ 
		150k                                                                      & 49.5  & 94.24    &  349.04                                                         & 25.88  & 108.5    & 318.04                                                     \\ 
		200k                                                                      & 80.62  & 148.8  &  663.74                                                        & 77.14  & 181.06    & 827.42 \\ \hline 
	\end{tabular}
	\vspace{-0.2cm}
\end{table}

\begin{figure*}
    \centering
	\captionsetup{justification=centering}
    \stackunder[5pt]{\includegraphics[height=2.5cm]{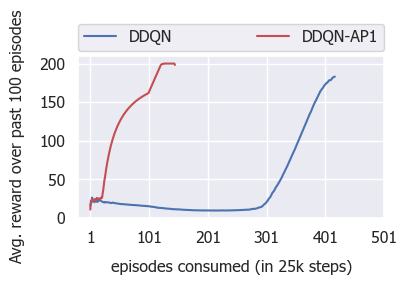}}{}%
    \stackunder[5pt]{\includegraphics[height=2.5cm]{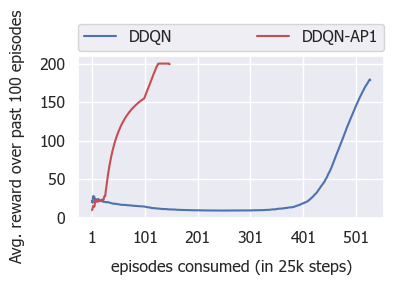}}{}%
	\stackunder[5pt]{\includegraphics[height=2.5cm]{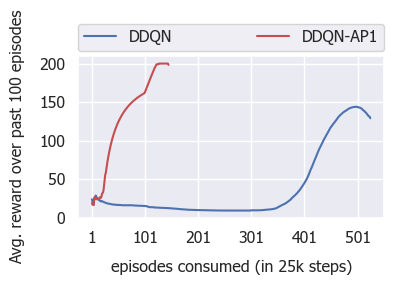}}{}
	\stackunder[5pt]{\includegraphics[height=2.5cm]{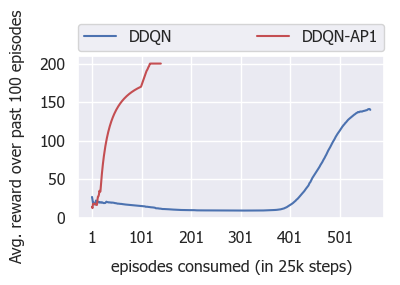}}{}%
	\stackunder[5pt]{\includegraphics[height=2.5cm]{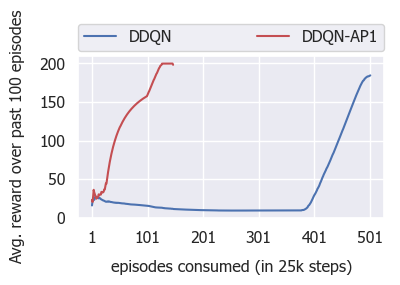}}{}%
	\label{task_example}
	\vspace{-0.1cm}
	\caption{\small Avg. reward over past 100 episodes for DDQN and DDQN-AP1 on cart-pole balancing over five training experiments with different random seeds. 2k exploration steps were used in training.}
	\vspace{-0.25cm}
\end{figure*}

Table 2 shows the test performance of the said algorithms recorded after 100k, 150k and 200k training steps. We see that DDQN-AP2 performs significantly better than DDQN (baseline) and DDQN-(AP1 +AP2) outperforms them both by a large margin. Note, the average scores for all three algorithms for 100k and 150k training steps are higher for the 30k exploration steps experiment than for the 60k exploration steps one.
This is because with only 30k exploration steps, the algorithms gets a longer training (post-exploration) phase than with 60k exploration steps. E.g., considering the 150k training steps, the learning phase has 120k steps for the 30k exploration steps experiment but only 90k steps for the 60k exploration steps one. However, at the 200k training step, DDQN-AP2, DDQN-(AP1+AP2) in the 60k exploration steps experiment performs significantly better than that for the 30k exploration steps one as the sufficiently longer exploration phase introduces more stability in learning and assists AP-based guidance in accelerating the training in the post-exploration/training phase.

\begin{wrapfigure}{r}{0.14\textwidth}
    \vspace{-0.8cm}
    \centering
    \includegraphics[height=2.5cm]{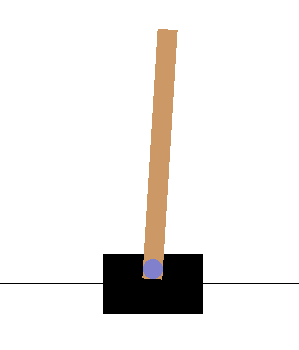}%
	\label{task_example}
	\vspace{-0.4cm}
	\caption{\small Snapshot of Cart-pole Balancing simulator.}
	\vspace{-0.3cm}
\end{wrapfigure}

\subsection{Cart-Pole Balancing} 
The Cart-Pole Balancing task\footnote{gym.openai.com} (see Appendix for details) is about learning to balance a pole attached to a cart which moves along a frictionless track (see Figure 8). The goal is to prevent it from falling  over by increasing and reducing the cart's velocity, where the allowed actions are \{\textit{push left}, \textit{push right}\} for 200 steps (episode length). Since the action space is discrete, we use DDQN. 

\textbf{AP Function.} Let $\rho_t$ be the pole's \textit{angular direction} (wrt the vertical axis Y of oscillation) at state $s_t$, defined as 
\vspace{-0.2cm}
\begin{equation}\label{key}
\rho_t = \begin{cases}
~\sign(~\theta_t~) & \text{if~ $\abs{~\theta_t~} \geq \Delta\theta_s$}\\
~0 & \text{Otherwise}
\end{cases}
\end{equation}
where $\theta_t$ is the pole angle (wrt Y), $\Delta\theta_s > 0$ is a \textit{tolerance value} (we empirically set it as 0.05) of the pole angle and $\sign(~\theta_t~) \in \{-1, +1\}$ denotes the sign of the pole angle at state $s_t$. $\rho_t$ basically denotes the side the pole is tilted due to the movement of the cart at a given state $s_t$, which is explained as follows: If $\abs{~\theta_t~} < \Delta\theta_s$, i.e., the pole angle is within a bound of $\Delta\theta_s$ around the axis Y, we consider that the pole is (almost) in a upright position (with negligible tilt). In such a case $\rho_t = 0$ and any action taken by the agent that causes the pole to stay in this bound is considered as permissible. When $\abs{~\theta_t~} \geq \Delta\theta_s$, the pole either tilts on the right side of the axis-Y with angle $\theta_t > \Delta\theta_s > \ang{0}$ ($\rho_t$ = 1) or on left side of the axis-Y with angle $\theta_t < -\Delta\theta_s < \ang{0}$ ($\rho_t$ = -1). In such situations, the pole is more vulnerable to fall over (if appropriate action is not chosen by the agent to recover it back). 

Let's denote the change in pole's angular velocity from state $s_t$ to $s_{t+1}$ is $\Delta v_{t,t+1}$ (i.e., $~|v_{t+1}|-|v_{t}|=\Delta v_{t,t+1}$) and change in pole's angular distance (wrt axis Y) from state $s_t$ to $s_{t+1}$ is $\Delta\theta_{t,t+1}$ (i.e., $~|\theta_{t+1}| - |\theta_{t}|=\Delta\theta_{t,t+1}$). Then, given the notion of $\rho_t$ defined in eqn. 8 (above), we define an type-1 AP function for the concerned cart-pole balancing problem as given by function $f_1$ below:
\vspace{-0.2cm}
\begin{equation}\label{key}
f_1(a_t, s_t) = \begin{cases}
0 & \text{if ~$\rho_{t+1} = \pm~1$, $~\Delta v_{t,t+1} > 0$, $~\Delta\theta_{t,t+1} > 0$}\\
1 & \text{Otherwise}
\end{cases}
\vspace{-0.05cm}
\end{equation}
$f_1$ is explained as follows: If due to the execution of action $a_t$ in $s_t$, $\rho_{t+1} = \pm~1$ in $s_{t+1}$ and both pole velocity and pole's angular distance increases at $s_{t+1}$ compared to that in $s_t$ ($~\Delta v_{t,t+1} > 0$, $~\Delta\theta_{t,t+1} > 0$), then $a_t$ in $s_t$ is not permissible, otherwise it is permissible.

\begin{wrapfigure}{r}{0.5\linewidth}
    \centering
    \vspace{-0.25cm}
    \includegraphics[height=2.93cm]{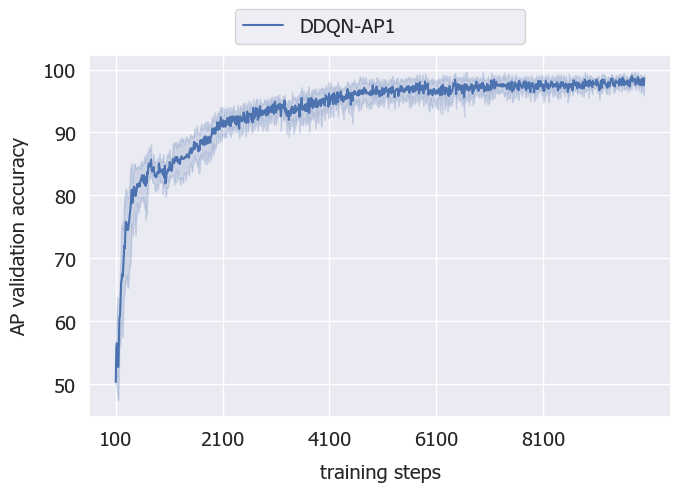}
	\label{task_example}
	\vspace{-0.4cm}
	\caption{Validation accuracy of AP1 predictor/classifier over training steps for cart-pole task.}
	\vspace{0.3cm}
	\scriptsize
	\centering
	\captionof{table}[]{\small Average test scores over 500 episodes of DDQN, DDQN-AP1 on cart-pole balancing. The 500 episodes consists of five test experiments conducted with five different random seeds (100 test episodes per test experiment).}
	\begin{tabular}{|l|c|c|c|c|}
    \hline
    \multicolumn{1}{|c|}{train. steps} & 5k              & 10k             & 15k            & 25k            \\ \hline
    DDQN                                 & 92.04           & 162.35          & 178.00         & 165.57         \\ \hline
    DDQN-AP1    & \textbf{199.72} & \textbf{200.0} & \textbf{200.0} & \textbf{200.0} \\ \hline
    \end{tabular}
    \vspace{-0.25cm}
\end{wrapfigure}

Note that, the pole starts in safe zone\footnote{Starting state of Cart-pole:
All observations are assigned a uniform random value between $\pm0.05$.} and the goal of the agent is to vertically align it with axis Y with pole velocity close to $0$ (minimize oscillation).  When $\rho_{t+1} = \pm~1$, then either because of $a_t$, pole may enter into danger zone ($\abs{~\theta_t~} \geq \Delta\theta_s$) from safe zone ($\abs{~\theta_t~} < \Delta\theta_s$) while transitioning from $s_t$ into $s_{t+1}$ or may remain in danger zone in both $s_t$ and $s_{t+1}$. In either case, if pole tip velocity and pole's angular distance in $s_{t+1}$ increases due to $a_t$, it implies $a_t$ makes the pole to fall over more in the same direction compared to that it was before (in $s_t$) and thus, $a_t$ is a non-permissible action for the balancing task. It is hard to mathematically prove that additional actions (forces applied in the same direction) will not recover the pole from the danger zone as we do not know the internal working of the simulation system. However, our AP1 function works well and it achieves the perfect test score (in the paper), which shows that it does not cut off too much exploration space and leave a sub-optimal solution. 

Designing an AP2 function is difficult for the cart-pole balancing task. One strategy for AP2 would be to push the cart right when the pole is tilted right ($\rho_{t+1} = 1$) and vice-versa. However, if we apply consecutive right pushes, the pole gain upward momentum (due to increase in angular velocity) and eventually, tilts to the left side. Similarly, consecutive left pushes when the pole is tilted left eventually makes the pole tilt on right side. Such policy leads to oscillating behavior of the pole and eventually terminates the episode (when pole angle becomes $> \ang{12}$). Our experiments confirmed this. Thus, designing type-2 AP function (knowing good and bad policies beforehand at a state) is hard for this task compared to type-1 AP function.

\textbf{Results.} In the Cart-pole task, the agent receives the same reward (1.0) for any action (permissible or non-permissible) executed on the environment (see Supplementary). Thus, we use virtual stopping (VS) strategy in DDQN-AP1.

We trained DDQN and DDQN-AP1 for 25k steps with $\epsilon$-greedy strategy and observed drastic growth in reward during training for DDQN-AP1 compared to that for DDQN. The learning curves of DDQN and DDQN-AP1 in Fig. 9 show the usefulness of formulated AP1 function (Eqn. 9) and virtual stopping in accelerating the RL training. We also note that AP predictor's validation accuracy always stays above 90\% during training and stabilizes with an average of 95.8\% as shown in Fig. 10. 

Next, we evaluate the performance of the trained DDQN-AP1 and DDQN in terms of the average score over 500 test episodes. Table 3 shows the test results of the algorithms recorded after 5k, 10k, 15k and 25k training steps (200 is the perfect score). We see that DDQN-AP1 not only outperforms DDQN, but also learns much quicker to solve the task compared to DDQN with less steps (learned to solve in just 5k training steps).

\section{Limitations and Future Work}
Our work deals with a class of RL problems with the SAP property. Examples of such problems \textit{primarily} include robot navigation, planning for solving a task, games, etc.  For such a problem, it is often not hard to specify an AP function with expert's domain knowledge. However, by no means do we claim that the SAP property is applicable to all RL problems. Some RL problems don't have SAP property or it is hard to specify AP functions for them (e.g., environments with high dimensional action space like humanoid (openai gym)). We should also note that this work does not focus on designing procedures to identify AP functions as AP functions depend on specific applications. Also, it does not require the user to provide an ``\textit{optimal AP function}" as several AP functions can usually be designed for a given problem. This work mainly aims to provide a framework to guide the existing Deep RL algorithms when a fairly good AP function can be designed for an environment. In the future, we plan to extend our framework to multi-dimensional continuous and discrete action spaces and apply it to other applications. We also plan to investigate the possibility of learning the AP function automatically for an application based on some exploration steps.

\section{Conclusion}
This paper proposed an novel property, called state-action permissibilty (SAP), for speeding up RL training in problems with this property. To leverage this property, we proposed two types of action permissibility and designed a framework for AP-guided exploration to help Deep RL algorithms select promising actions in each state in training together with a virtual-stopping strategy. Experimental results showed that the proposed method is highly effective.

\bibliographystyle{ACM-Reference-Format}
\bibliography{acmart}


\begin{thebibliography}{43}


\ifx \showCODEN    \undefined \def \showCODEN     #1{\unskip}     \fi
\ifx \showDOI      \undefined \def \showDOI       #1{#1}\fi
\ifx \showISBNx    \undefined \def \showISBNx     #1{\unskip}     \fi
\ifx \showISBNxiii \undefined \def \showISBNxiii  #1{\unskip}     \fi
\ifx \showISSN     \undefined \def \showISSN      #1{\unskip}     \fi
\ifx \showLCCN     \undefined \def \showLCCN      #1{\unskip}     \fi
\ifx \shownote     \undefined \def \shownote      #1{#1}          \fi
\ifx \showarticletitle \undefined \def \showarticletitle #1{#1}   \fi
\ifx \showURL      \undefined \def \showURL       {\relax}        \fi
\providecommand\bibfield[2]{#2}
\providecommand\bibinfo[2]{#2}
\providecommand\natexlab[1]{#1}
\providecommand\showeprint[2][]{arXiv:#2}

\bibitem[\protect\citeauthoryear{Abel, Hershkowitz, Barth-Maron, Brawner,
  O'Farrell, MacGlashan, and Tellex}{Abel et~al\mbox{.}}{2015}]%
        {abel2015goal}
\bibfield{author}{\bibinfo{person}{David Abel}, \bibinfo{person}{David~Ellis
  Hershkowitz}, \bibinfo{person}{Gabriel Barth-Maron}, \bibinfo{person}{Stephen
  Brawner}, \bibinfo{person}{Kevin O'Farrell}, \bibinfo{person}{James
  MacGlashan}, {and} \bibinfo{person}{Stefanie Tellex}.}
  \bibinfo{year}{2015}\natexlab{}.
\newblock \showarticletitle{Goal-Based Action Priors}. In
  \bibinfo{booktitle}{\emph{International Conference on Automated Planning and
  Scheduling}}.
\newblock


\bibitem[\protect\citeauthoryear{Asmuth, Littman, and Zinkov}{Asmuth
  et~al\mbox{.}}{2008}]%
        {asmuth2008potential}
\bibfield{author}{\bibinfo{person}{John Asmuth}, \bibinfo{person}{Michael~L
  Littman}, {and} \bibinfo{person}{Robert Zinkov}.}
  \bibinfo{year}{2008}\natexlab{}.
\newblock \showarticletitle{Potential-based Shaping in Model-based
  Reinforcement Learning.}. In \bibinfo{booktitle}{\emph{AAAI}}.
\newblock


\bibitem[\protect\citeauthoryear{Bacon, Harb, and Precup}{Bacon
  et~al\mbox{.}}{2017}]%
        {bacon2017option}
\bibfield{author}{\bibinfo{person}{Pierre-Luc Bacon}, \bibinfo{person}{Jean
  Harb}, {and} \bibinfo{person}{Doina Precup}.}
  \bibinfo{year}{2017}\natexlab{}.
\newblock \showarticletitle{The Option-Critic Architecture.}. In
  \bibinfo{booktitle}{\emph{AAAI}}.
\newblock


\bibitem[\protect\citeauthoryear{Bai and Russell}{Bai and Russell}{2017}]%
        {bai2017efficient}
\bibfield{author}{\bibinfo{person}{Aijun Bai} {and} \bibinfo{person}{Stuart
  Russell}.} \bibinfo{year}{2017}\natexlab{}.
\newblock \showarticletitle{Efficient reinforcement learning with hierarchies
  of machines by leveraging internal transitions}. In
  \bibinfo{booktitle}{\emph{IJCAI}}.
\newblock


\bibitem[\protect\citeauthoryear{Berkenkamp, Turchetta, Schoellig, and
  Krause}{Berkenkamp et~al\mbox{.}}{2017}]%
        {berkenkamp2017safe}
\bibfield{author}{\bibinfo{person}{Felix Berkenkamp}, \bibinfo{person}{Matteo
  Turchetta}, \bibinfo{person}{Angela Schoellig}, {and}
  \bibinfo{person}{Andreas Krause}.} \bibinfo{year}{2017}\natexlab{}.
\newblock \showarticletitle{Safe model-based reinforcement learning with
  stability guarantees}. In \bibinfo{booktitle}{\emph{NIPS}}.
\newblock


\bibitem[\protect\citeauthoryear{Clavera, Rothfuss, Schulman, Fujita, Asfour,
  and Abbeel}{Clavera et~al\mbox{.}}{2018}]%
        {clavera2018model}
\bibfield{author}{\bibinfo{person}{Ignasi Clavera}, \bibinfo{person}{Jonas
  Rothfuss}, \bibinfo{person}{John Schulman}, \bibinfo{person}{Yasuhiro
  Fujita}, \bibinfo{person}{Tamim Asfour}, {and} \bibinfo{person}{Pieter
  Abbeel}.} \bibinfo{year}{2018}\natexlab{}.
\newblock \showarticletitle{Model-based reinforcement learning via meta-policy
  optimization}.
\newblock \bibinfo{journal}{\emph{arXiv preprint arXiv:1809.05214}}
  (\bibinfo{year}{2018}).
\newblock


\bibitem[\protect\citeauthoryear{Dalal, Dvijotham, Vecerik, Hester, Paduraru,
  and Tassa}{Dalal et~al\mbox{.}}{2018}]%
        {dalal2018safe}
\bibfield{author}{\bibinfo{person}{Gal Dalal}, \bibinfo{person}{Krishnamurthy
  Dvijotham}, \bibinfo{person}{Matej Vecerik}, \bibinfo{person}{Todd Hester},
  \bibinfo{person}{Cosmin Paduraru}, {and} \bibinfo{person}{Yuval Tassa}.}
  \bibinfo{year}{2018}\natexlab{}.
\newblock \showarticletitle{Safe exploration in continuous action spaces}.
\newblock \bibinfo{journal}{\emph{arXiv preprint arXiv:1801.08757}}
  (\bibinfo{year}{2018}).
\newblock


\bibitem[\protect\citeauthoryear{Deisenroth and Rasmussen}{Deisenroth and
  Rasmussen}{2011}]%
        {deisenroth2011pilco}
\bibfield{author}{\bibinfo{person}{Marc Deisenroth} {and}
  \bibinfo{person}{Carl~E Rasmussen}.} \bibinfo{year}{2011}\natexlab{}.
\newblock \showarticletitle{PILCO: A model-based and data-efficient approach to
  policy search}. In \bibinfo{booktitle}{\emph{ICML}}.
\newblock


\bibitem[\protect\citeauthoryear{Duan, Schulman, Chen, Bartlett, Sutskever, and
  Abbeel}{Duan et~al\mbox{.}}{2016}]%
        {duan2016rl}
\bibfield{author}{\bibinfo{person}{Yan Duan}, \bibinfo{person}{John Schulman},
  \bibinfo{person}{Xi Chen}, \bibinfo{person}{Peter~L Bartlett},
  \bibinfo{person}{Ilya Sutskever}, {and} \bibinfo{person}{Pieter Abbeel}.}
  \bibinfo{year}{2016}\natexlab{}.
\newblock \showarticletitle{RL $^2$: Fast Reinforcement Learning via Slow
  Reinforcement Learning}. In \bibinfo{booktitle}{\emph{arXiv preprint
  arXiv:1611.02779}}.
\newblock


\bibitem[\protect\citeauthoryear{Dubey, Agrawal, Pathak, Griffiths~L., and
  Efros}{Dubey et~al\mbox{.}}{2018}]%
        {Dubey2018}
\bibfield{author}{\bibinfo{person}{Rachit Dubey}, \bibinfo{person}{Pulkit
  Agrawal}, \bibinfo{person}{Deepak Pathak}, \bibinfo{person}{Thomas
  Griffiths~L.}, {and} \bibinfo{person}{Alexei~A. Efros}.}
  \bibinfo{year}{2018}\natexlab{}.
\newblock \showarticletitle{Investigating Human Priors for Playing Video
  Games}. In \bibinfo{booktitle}{\emph{ICML}}.
\newblock


\bibitem[\protect\citeauthoryear{Even-Dar and Mansour}{Even-Dar and
  Mansour}{2003}]%
        {even2003learning}
\bibfield{author}{\bibinfo{person}{Eyal Even-Dar} {and} \bibinfo{person}{Yishay
  Mansour}.} \bibinfo{year}{2003}\natexlab{}.
\newblock \showarticletitle{Learning rates for Q-learning}.
\newblock \bibinfo{journal}{\emph{JMLR}} (\bibinfo{year}{2003}).
\newblock


\bibitem[\protect\citeauthoryear{Finn, Abbeel, and Levine}{Finn
  et~al\mbox{.}}{2017}]%
        {finn2017model}
\bibfield{author}{\bibinfo{person}{Chelsea Finn}, \bibinfo{person}{Pieter
  Abbeel}, {and} \bibinfo{person}{Sergey Levine}.}
  \bibinfo{year}{2017}\natexlab{}.
\newblock \showarticletitle{Model-agnostic meta-learning for fast adaptation of
  deep networks}.
\newblock \bibinfo{journal}{\emph{ICML}} (\bibinfo{year}{2017}).
\newblock


\bibitem[\protect\citeauthoryear{Fulda, Ricks, Murdoch, and Wingate}{Fulda
  et~al\mbox{.}}{2017}]%
        {fulda2017can}
\bibfield{author}{\bibinfo{person}{Nancy Fulda}, \bibinfo{person}{Daniel
  Ricks}, \bibinfo{person}{Ben Murdoch}, {and} \bibinfo{person}{David
  Wingate}.} \bibinfo{year}{2017}\natexlab{}.
\newblock \showarticletitle{What can you do with a rock? affordance extraction
  via word embeddings}.
\newblock \bibinfo{journal}{\emph{arXiv preprint arXiv:1703.03429}}
  (\bibinfo{year}{2017}).
\newblock


\bibitem[\protect\citeauthoryear{Gao and Toni}{Gao and Toni}{2015}]%
        {gao2015potential}
\bibfield{author}{\bibinfo{person}{Yang Gao} {and} \bibinfo{person}{Francesca
  Toni}.} \bibinfo{year}{2015}\natexlab{}.
\newblock \showarticletitle{Potential Based Reward Shaping for Hierarchical
  Reinforcement Learning.}. In \bibinfo{booktitle}{\emph{IJCAI}}.
\newblock


\bibitem[\protect\citeauthoryear{Girgin, Polat, and Alhajj}{Girgin
  et~al\mbox{.}}{2010}]%
        {girgin2010improving}
\bibfield{author}{\bibinfo{person}{Sertan Girgin}, \bibinfo{person}{Faruk
  Polat}, {and} \bibinfo{person}{Reda Alhajj}.}
  \bibinfo{year}{2010}\natexlab{}.
\newblock \showarticletitle{Improving reinforcement learning by using sequence
  trees}.
\newblock \bibinfo{journal}{\emph{Machine Learning}} (\bibinfo{year}{2010}).
\newblock


\bibitem[\protect\citeauthoryear{Kamalapurkar, Walters, and Dixon}{Kamalapurkar
  et~al\mbox{.}}{2016}]%
        {kamalapurkar2016model}
\bibfield{author}{\bibinfo{person}{Rushikesh Kamalapurkar},
  \bibinfo{person}{Patrick Walters}, {and} \bibinfo{person}{Warren~E Dixon}.}
  \bibinfo{year}{2016}\natexlab{}.
\newblock \showarticletitle{Model-based reinforcement learning for approximate
  optimal regulation}. In \bibinfo{booktitle}{\emph{Automatica}}.
\newblock


\bibitem[\protect\citeauthoryear{Kohl and Stone}{Kohl and Stone}{2004}]%
        {kohl2004policy}
\bibfield{author}{\bibinfo{person}{Nate Kohl} {and} \bibinfo{person}{Peter
  Stone}.} \bibinfo{year}{2004}\natexlab{}.
\newblock \showarticletitle{Policy gradient reinforcement learning for fast
  quadrupedal locomotion}. In \bibinfo{booktitle}{\emph{ICRA}}.
\newblock


\bibitem[\protect\citeauthoryear{Kulkarni, Narasimhan, Saeedi, and
  Tenenbaum}{Kulkarni et~al\mbox{.}}{2016}]%
        {kulkarni2016hierarchical}
\bibfield{author}{\bibinfo{person}{Tejas~D Kulkarni}, \bibinfo{person}{Karthik
  Narasimhan}, \bibinfo{person}{Ardavan Saeedi}, {and} \bibinfo{person}{Josh
  Tenenbaum}.} \bibinfo{year}{2016}\natexlab{}.
\newblock \showarticletitle{Hierarchical deep reinforcement learning:
  Integrating temporal abstraction and intrinsic motivation}. In
  \bibinfo{booktitle}{\emph{NIPS}}.
\newblock


\bibitem[\protect\citeauthoryear{Lillicrap, Hunt, Pritzel, Heess, Erez, Tassa,
  Silver, and Wierstra}{Lillicrap et~al\mbox{.}}{2016}]%
        {lillicrap2015continuous}
\bibfield{author}{\bibinfo{person}{Timothy~P Lillicrap},
  \bibinfo{person}{Jonathan~J Hunt}, \bibinfo{person}{Alexander Pritzel},
  \bibinfo{person}{Nicolas Heess}, \bibinfo{person}{Tom Erez},
  \bibinfo{person}{Yuval Tassa}, \bibinfo{person}{David Silver}, {and}
  \bibinfo{person}{Daan Wierstra}.} \bibinfo{year}{2016}\natexlab{}.
\newblock \showarticletitle{Continuous control with deep reinforcement
  learning}. In \bibinfo{booktitle}{\emph{ICLR}}.
\newblock


\bibitem[\protect\citeauthoryear{Lipton, Li, Gao, Li, Ahmed, and Deng}{Lipton
  et~al\mbox{.}}{2018}]%
        {lipton2018bbq}
\bibfield{author}{\bibinfo{person}{Zachary Lipton}, \bibinfo{person}{Xiujun
  Li}, \bibinfo{person}{Jianfeng Gao}, \bibinfo{person}{Lihong Li},
  \bibinfo{person}{Faisal Ahmed}, {and} \bibinfo{person}{Li Deng}.}
  \bibinfo{year}{2018}\natexlab{}.
\newblock \showarticletitle{Bbq-networks: Efficient exploration in deep
  reinforcement learning for task-oriented dialogue systems}. In
  \bibinfo{booktitle}{\emph{AAAI}}.
\newblock


\bibitem[\protect\citeauthoryear{Lipton, Azizzadenesheli, Kumar, Li, Gao, and
  Deng}{Lipton et~al\mbox{.}}{2016}]%
        {lipton2016combating}
\bibfield{author}{\bibinfo{person}{Zachary~C Lipton}, \bibinfo{person}{Kamyar
  Azizzadenesheli}, \bibinfo{person}{Abhishek Kumar}, \bibinfo{person}{Lihong
  Li}, \bibinfo{person}{Jianfeng Gao}, {and} \bibinfo{person}{Li Deng}.}
  \bibinfo{year}{2016}\natexlab{}.
\newblock \showarticletitle{Combating reinforcement learning's sisyphean curse
  with intrinsic fear}.
\newblock \bibinfo{journal}{\emph{arXiv preprint arXiv:1611.01211}}
  (\bibinfo{year}{2016}).
\newblock


\bibitem[\protect\citeauthoryear{Loiacono, Cardamone, and Lanzi}{Loiacono
  et~al\mbox{.}}{2013}]%
        {loiacono2013simulated}
\bibfield{author}{\bibinfo{person}{Daniele Loiacono}, \bibinfo{person}{Luigi
  Cardamone}, {and} \bibinfo{person}{Pier~Luca Lanzi}.}
  \bibinfo{year}{2013}\natexlab{}.
\newblock \showarticletitle{Simulated car racing championship: Competition
  software manual}.
\newblock \bibinfo{journal}{\emph{arXiv preprint arXiv:1304.1672}}
  (\bibinfo{year}{2013}).
\newblock


\bibitem[\protect\citeauthoryear{Mahajan and Tulabandhula}{Mahajan and
  Tulabandhula}{2017}]%
        {mahajan2017symmetry}
\bibfield{author}{\bibinfo{person}{Anuj Mahajan} {and} \bibinfo{person}{Theja
  Tulabandhula}.} \bibinfo{year}{2017}\natexlab{}.
\newblock \showarticletitle{Symmetry Detection and Exploitation for Function
  Approximation in Deep RL}. In \bibinfo{booktitle}{\emph{AAMAS}}.
\newblock


\bibitem[\protect\citeauthoryear{Marom and Rosman}{Marom and Rosman}{2018}]%
        {marom2018belief}
\bibfield{author}{\bibinfo{person}{Ofir Marom} {and} \bibinfo{person}{Benjamin
  Rosman}.} \bibinfo{year}{2018}\natexlab{}.
\newblock \showarticletitle{Belief reward shaping in reinforcement learning}.
  In \bibinfo{booktitle}{\emph{AAAI}}.
\newblock


\bibitem[\protect\citeauthoryear{Mazumder, Liu, Wang, Zhu, Liu, and
  Li}{Mazumder et~al\mbox{.}}{2018}]%
        {mazumder2018action}
\bibfield{author}{\bibinfo{person}{Sahisnu Mazumder}, \bibinfo{person}{Bing
  Liu}, \bibinfo{person}{Shuai Wang}, \bibinfo{person}{Yingxuan Zhu},
  \bibinfo{person}{Lifeng Liu}, {and} \bibinfo{person}{Jian Li}.}
  \bibinfo{year}{2018}\natexlab{}.
\newblock \showarticletitle{Action permissibility in deep reinforcement
  learning and application to autonomous driving}. In
  \bibinfo{booktitle}{\emph{2018 SIGKDD Conference on Knowledge Discovery and
  Data Mining (KDD)}}.
\newblock


\bibitem[\protect\citeauthoryear{Mnih, Kavukcuoglu, Silver, Graves, Antonoglou,
  Wierstra, and Riedmiller}{Mnih et~al\mbox{.}}{2013}]%
        {mnih2013playing}
\bibfield{author}{\bibinfo{person}{Volodymyr Mnih}, \bibinfo{person}{Koray
  Kavukcuoglu}, \bibinfo{person}{David Silver}, \bibinfo{person}{Alex Graves},
  \bibinfo{person}{Ioannis Antonoglou}, \bibinfo{person}{Daan Wierstra}, {and}
  \bibinfo{person}{Martin Riedmiller}.} \bibinfo{year}{2013}\natexlab{}.
\newblock \showarticletitle{Playing atari with deep reinforcement learning}.
\newblock \bibinfo{journal}{\emph{arXiv preprint arXiv:1312.5602}}
  (\bibinfo{year}{2013}).
\newblock


\bibitem[\protect\citeauthoryear{Mnih, Kavukcuoglu, Silver, Rusu, Veness,
  Bellemare, Graves, Riedmiller, Fidjeland, Ostrovski, et~al\mbox{.}}{Mnih
  et~al\mbox{.}}{2015}]%
        {mnih2015human}
\bibfield{author}{\bibinfo{person}{Volodymyr Mnih}, \bibinfo{person}{Koray
  Kavukcuoglu}, \bibinfo{person}{David Silver}, \bibinfo{person}{Andrei~A
  Rusu}, \bibinfo{person}{Joel Veness}, \bibinfo{person}{Marc~G Bellemare},
  \bibinfo{person}{Alex Graves}, \bibinfo{person}{Martin Riedmiller},
  \bibinfo{person}{Andreas~K Fidjeland}, \bibinfo{person}{Georg Ostrovski},
  {et~al\mbox{.}}} \bibinfo{year}{2015}\natexlab{}.
\newblock \showarticletitle{Human-level control through deep reinforcement
  learning}.
\newblock \bibinfo{journal}{\emph{Nature}} (\bibinfo{year}{2015}).
\newblock


\bibitem[\protect\citeauthoryear{Nagabandi, Kahn, Fearing, and
  Levine}{Nagabandi et~al\mbox{.}}{2018}]%
        {nagabandi2018neural}
\bibfield{author}{\bibinfo{person}{Anusha Nagabandi}, \bibinfo{person}{Gregory
  Kahn}, \bibinfo{person}{Ronald~S Fearing}, {and} \bibinfo{person}{Sergey
  Levine}.} \bibinfo{year}{2018}\natexlab{}.
\newblock \showarticletitle{Neural network dynamics for model-based deep
  reinforcement learning with model-free fine-tuning}. In
  \bibinfo{booktitle}{\emph{ICRA}}.
\newblock


\bibitem[\protect\citeauthoryear{Nair, McGrew, Andrychowicz, Zaremba, and
  Abbeel}{Nair et~al\mbox{.}}{2017}]%
        {nair2017overcoming}
\bibfield{author}{\bibinfo{person}{Ashvin Nair}, \bibinfo{person}{Bob McGrew},
  \bibinfo{person}{Marcin Andrychowicz}, \bibinfo{person}{Wojciech Zaremba},
  {and} \bibinfo{person}{Pieter Abbeel}.} \bibinfo{year}{2017}\natexlab{}.
\newblock \showarticletitle{Overcoming exploration in reinforcement learning
  with demonstrations}.
\newblock  (\bibinfo{year}{2017}).
\newblock


\bibitem[\protect\citeauthoryear{Narendra, Wang, and Mukhopadhay}{Narendra
  et~al\mbox{.}}{2016}]%
        {narendra2016fast}
\bibfield{author}{\bibinfo{person}{Kumpati~S Narendra}, \bibinfo{person}{Yu
  Wang}, {and} \bibinfo{person}{Snehasis Mukhopadhay}.}
  \bibinfo{year}{2016}\natexlab{}.
\newblock \showarticletitle{Fast Reinforcement Learning using multiple models}.
  In \bibinfo{booktitle}{\emph{CDC}}.
\newblock


\bibitem[\protect\citeauthoryear{Ng, Harada, and Russell}{Ng
  et~al\mbox{.}}{1999}]%
        {ng1999policy}
\bibfield{author}{\bibinfo{person}{Andrew~Y Ng}, \bibinfo{person}{Daishi
  Harada}, {and} \bibinfo{person}{Stuart Russell}.}
  \bibinfo{year}{1999}\natexlab{}.
\newblock \showarticletitle{Policy invariance under reward transformations:
  Theory and application to reward shaping}. In
  \bibinfo{booktitle}{\emph{ICML}}.
\newblock


\bibitem[\protect\citeauthoryear{Osband, Russo, and Van~Roy}{Osband
  et~al\mbox{.}}{2013}]%
        {osband2013more}
\bibfield{author}{\bibinfo{person}{Ian Osband}, \bibinfo{person}{Daniel Russo},
  {and} \bibinfo{person}{Benjamin Van~Roy}.} \bibinfo{year}{2013}\natexlab{}.
\newblock \showarticletitle{(More) efficient reinforcement learning via
  posterior sampling}. In \bibinfo{booktitle}{\emph{NIPS}}.
\newblock


\bibitem[\protect\citeauthoryear{Rosman and Ramamoorthy}{Rosman and
  Ramamoorthy}{2012}]%
        {rosman2012good}
\bibfield{author}{\bibinfo{person}{Benjamin Rosman} {and}
  \bibinfo{person}{Subramanian Ramamoorthy}.} \bibinfo{year}{2012}\natexlab{}.
\newblock \showarticletitle{What good are actions? Accelerating learning using
  learned action priors}. In \bibinfo{booktitle}{\emph{ICDL}}.
\newblock


\bibitem[\protect\citeauthoryear{Sallab, Abdou, Perot, and Yogamani}{Sallab
  et~al\mbox{.}}{2016}]%
        {sallab2016end}
\bibfield{author}{\bibinfo{person}{Ahmad~El Sallab}, \bibinfo{person}{Mohammed
  Abdou}, \bibinfo{person}{Etienne Perot}, {and} \bibinfo{person}{Senthil
  Yogamani}.} \bibinfo{year}{2016}\natexlab{}.
\newblock \showarticletitle{End-to-End Deep Reinforcement Learning for Lane
  Keeping Assist}.
\newblock \bibinfo{journal}{\emph{arXiv preprint arXiv:1612.04340}}
  (\bibinfo{year}{2016}).
\newblock


\bibitem[\protect\citeauthoryear{Sallab, Abdou, Perot, and Yogamani}{Sallab
  et~al\mbox{.}}{2017}]%
        {sallab2017deep}
\bibfield{author}{\bibinfo{person}{Ahmad~EL Sallab}, \bibinfo{person}{Mohammed
  Abdou}, \bibinfo{person}{Etienne Perot}, {and} \bibinfo{person}{Senthil
  Yogamani}.} \bibinfo{year}{2017}\natexlab{}.
\newblock \showarticletitle{Deep reinforcement learning framework for
  autonomous driving}.
\newblock \bibinfo{journal}{\emph{Electronic Imaging}} (\bibinfo{year}{2017}).
\newblock


\bibitem[\protect\citeauthoryear{Schwab and Ray}{Schwab and Ray}{2017}]%
        {schwab2017offline}
\bibfield{author}{\bibinfo{person}{Devin Schwab} {and} \bibinfo{person}{Soumya
  Ray}.} \bibinfo{year}{2017}\natexlab{}.
\newblock \showarticletitle{Offline reinforcement learning with task
  hierarchies}.
\newblock \bibinfo{journal}{\emph{Machine Learning}} (\bibinfo{year}{2017}).
\newblock


\bibitem[\protect\citeauthoryear{Sutton and Barto}{Sutton and Barto}{2017}]%
        {sutton2017reinforcement}
\bibfield{author}{\bibinfo{person}{Richard~S Sutton} {and}
  \bibinfo{person}{Andrew~G Barto}.} \bibinfo{year}{2017}\natexlab{}.
\newblock \bibinfo{booktitle}{\emph{Reinforcement learning: An introduction}}.
\newblock
  \bibinfo{publisher}{http://incompleteideas.net/book/bookdraft2017nov5.pdf}.
\newblock


\bibitem[\protect\citeauthoryear{Tsividis, Pouncy, Xu, Tenenbaum, and
  Gershman}{Tsividis et~al\mbox{.}}{2017}]%
        {Tsividis2017}
\bibfield{author}{\bibinfo{person}{Pedro~A. Tsividis}, \bibinfo{person}{Thomas
  Pouncy}, \bibinfo{person}{Jacqueline~L. Xu}, \bibinfo{person}{Joshua~B.
  Tenenbaum}, {and} \bibinfo{person}{Samuel~J. Gershman}.}
  \bibinfo{year}{2017}\natexlab{}.
\newblock \showarticletitle{Human Learning in Atari}. In
  \bibinfo{booktitle}{\emph{AAAI Symposium on Science of Intelligence}}.
\newblock


\bibitem[\protect\citeauthoryear{van Hasselt, Guez, and Silver}{van Hasselt
  et~al\mbox{.}}{2016}]%
        {van2016deep}
\bibfield{author}{\bibinfo{person}{Hado van Hasselt}, \bibinfo{person}{Arthur
  Guez}, {and} \bibinfo{person}{David Silver}.}
  \bibinfo{year}{2016}\natexlab{}.
\newblock \showarticletitle{Deep Reinforcement Learning with Double
  Q-Learning}. In \bibinfo{booktitle}{\emph{AAAI}}.
\newblock


\bibitem[\protect\citeauthoryear{Wang, Kurth-Nelson, Tirumala, Soyer, Leibo,
  Munos, Blundell, Kumaran, and Botvinick}{Wang et~al\mbox{.}}{2016}]%
        {wang2016learning}
\bibfield{author}{\bibinfo{person}{Jane~X Wang}, \bibinfo{person}{Zeb
  Kurth-Nelson}, \bibinfo{person}{Dhruva Tirumala}, \bibinfo{person}{Hubert
  Soyer}, \bibinfo{person}{Joel~Z Leibo}, \bibinfo{person}{Remi Munos},
  \bibinfo{person}{Charles Blundell}, \bibinfo{person}{Dharshan Kumaran}, {and}
  \bibinfo{person}{Matt Botvinick}.} \bibinfo{year}{2016}\natexlab{}.
\newblock \showarticletitle{Learning to reinforcement learn}.
\newblock \bibinfo{journal}{\emph{arXiv preprint arXiv:1611.05763}}
  (\bibinfo{year}{2016}).
\newblock


\bibitem[\protect\citeauthoryear{Watkins and Dayan}{Watkins and Dayan}{1992}]%
        {watkins1992q}
\bibfield{author}{\bibinfo{person}{Christopher~JCH Watkins} {and}
  \bibinfo{person}{Peter Dayan}.} \bibinfo{year}{1992}\natexlab{}.
\newblock \showarticletitle{Q-learning}. In \bibinfo{booktitle}{\emph{Machine
  learning}}.
\newblock


\bibitem[\protect\citeauthoryear{Wu, Guo, and Liu}{Wu et~al\mbox{.}}{2017}]%
        {wu2017adaptive}
\bibfield{author}{\bibinfo{person}{Huasen Wu}, \bibinfo{person}{Xueying Guo},
  {and} \bibinfo{person}{Xin Liu}.} \bibinfo{year}{2017}\natexlab{}.
\newblock \showarticletitle{Adaptive Exploration-Exploitation Tradeoff for
  Opportunistic Bandits}.
\newblock \bibinfo{journal}{\emph{arXiv preprint arXiv:1709.04004}}
  (\bibinfo{year}{2017}).
\newblock


\bibitem[\protect\citeauthoryear{Zahavy, Haroush, Merlis, Mankowitz, and
  Mannor}{Zahavy et~al\mbox{.}}{2018}]%
        {zahavy2018learn}
\bibfield{author}{\bibinfo{person}{Tom Zahavy}, \bibinfo{person}{Matan
  Haroush}, \bibinfo{person}{Nadav Merlis}, \bibinfo{person}{Daniel~J
  Mankowitz}, {and} \bibinfo{person}{Shie Mannor}.}
  \bibinfo{year}{2018}\natexlab{}.
\newblock \showarticletitle{Learn What Not to Learn: Action Elimination with
  Deep Reinforcement Learning}. In \bibinfo{booktitle}{\emph{NeurIPS}}.
\newblock


\end{thebibliography}

\appendix

\section{Lane Keeping: Description and Implementation Details} 

\textbf{The TORCS Simulator.} The Open Racing Car Simulator (TORCS) provides us with graphics and physics engines for Simulated Car Racing (SCR). The availability of the diverse set of road tracks with varying curvatures, landscapes and slopes in TORCS makes it an appropriate choice for model evaluation in different driving scenarios. It also allows us to play with different car control parameters like steering angle, velocity, acceleration, brakes, etc. 
More details can be found at  \url{https://www.cs.bgu.ac.il/~yakobis/files/patch_manual.pdf}. For our lane keeping (steering control) problem setting, we used five sensor variables that are sufficient for learning the steering control action as presented in Table 4. The \textit{trackPos} parameter in Table 4 has been used as a parameter for designing the AP1 function in our concerned lane keeping task (see Equation 5 in the paper).

\begin{table}[th]
	\scriptsize
	\centering
	\label{my-label}
	\vspace{-0.2cm}
	\begin{tabular}{ccccc}
		\textbf{Wheel-2}  &  \textbf{Spring} & \textbf{E-road} & \textbf{CG Track 3} & \textbf{Oleth Ross}\\
		(6205.46m)  &  (22129.77m) & (3260.43m) & (2843.10m) & (6282.81m) \\
		\includegraphics[height=0.93cm]{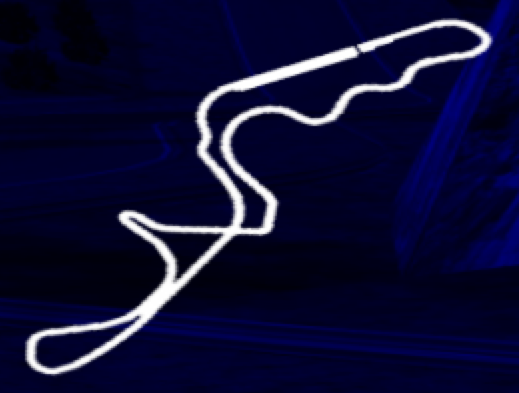} &	\includegraphics[height=0.93cm]{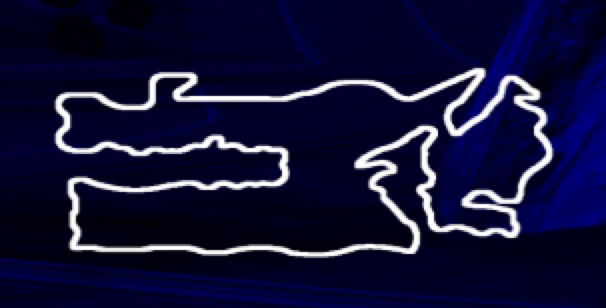} & \includegraphics[height=0.93cm]{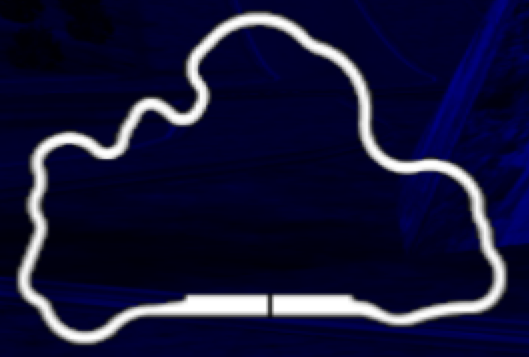} &	\includegraphics[height=0.93cm]{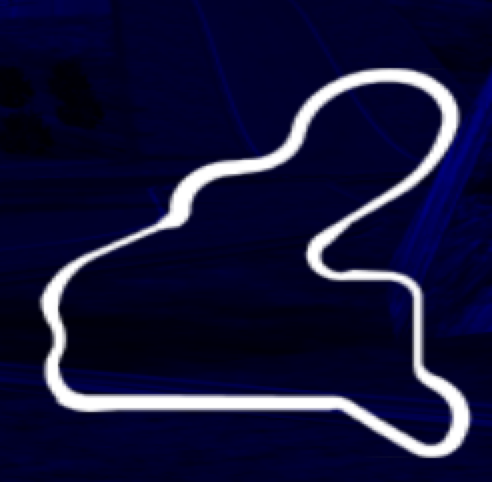} &
		\includegraphics[height=0.93cm]{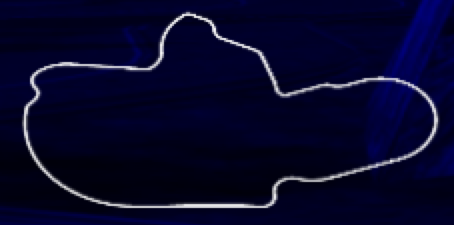}
	\end{tabular}
	\vspace{-0.15cm}
	\captionof{figure}{Various road tracks used in our experiments.}
	\vspace{-0.25cm}
\end{table}

Figure 11 shows the five tracks (with their lengths) used for evaluation. These tracks are diverse in landscapes and slopes. Among these 5 tracks, we used wheel-2 for training and the rest of the four tracks for testing. Due to various curvature variations, we consider wheel-2 as ideal for training all possible scenarios. In our evaluation setup, we deal with only steering control problem and use a gym torcs \textit{default} script for automatic speed control\footnote{\url{github.com/yanpanlau/DDPG-Keras-Torcs/blob/master/gym\_torcs.py}}. We use the default speed of the car as 100 and the max. value of acceleration (deceleration) applied at any given state is 0.2 (-0.2) for smooth speed transition (following a default settings of gym torcs simulator in ).

\begin{table}[t!]
	\scriptsize
	\centering
	\caption{\small TORCS state and action variables along with their descriptions used in our experiments.}
	\vspace{-0.1cm}
	\label{my-label}
	\begin{tabular}{|l|c|p{5.77cm}|}
		\hline
		\multicolumn{3}{|c|}{State Variables}                                                                                                                                                                                                                                                                                                                                                                                                                           \\ \hline
		\multicolumn{1}{|c|}{\textbf{Name}} & \multicolumn{1}{c|}{\textbf{Range (unit)}} & \multicolumn{1}{c|}{\textbf{Description}}                                                                                                                                                                                                                                                                                                                                     \\ \hline
		angle                               & {[}-$\pi$, $\pi${)} (rad)                      & Angle between the car direction and the direction of the track center axis.                                                                                                                                                                                                                                                                                                          \\ \hline
		trackPos                            & (-$\infty$,+$\infty$)                          & Distance between the car and the track center axis.  The value is  normalized w.r.t to the track width:  it is 0 when car is on the axis, -1 when the car is on the right edge of the track and +1 when it is on the left edge of the car. Values greater than 1 or smaller than -1 mean that the car is out of track.  \\ \hline
		speedX                              & (-$\infty$,+$\infty$)(km/h)                    & Speed of the car along its longitudinal axis.                                                                                                                                                                                                                                                                                              \\ \hline
		speedY                              & (-$\infty$,+$\infty$)(km/h)                    & Speed of the car along its transverse axis .                                                                                                                                                                                                                                                                                            \\ \hline
		speedZ                              & (-$\infty$,+$\infty$)(km/h)                    & Speed of the car along its Z-axis                                                                                                                                                                                                                                                                                                                   \\ \hline
		\multicolumn{3}{|c|}{Action}                                                                                                                                                                                                                                                                                                                                                                                                                                     \\ \hline
		Steering                      & {[}-1, 1{]}                                &  Steering value: -1 and +1 means respectively full right and  left, that corresponds to an angle of 0.366519 rad.                                                                                                                                                                                                                 \\ \hline
	\end{tabular}
	\vspace{-0.1cm}
	\normalsize
\end{table}

\textbf{Network Architecture.}  For our lane keeping (steering control) task, the Actor network is a feed forward (fully connected) network with 128 units in layer-1 and 256 units in layer-2 followed by the action projection (output) layer. In the Critic network, we first learn state representation $s$ using two fully connected layers of 128 and 256 units. We also learn a representation of the action $a$ chosen by Actor at state $s$ with one fully connected layer of 256 units.  Then, we concatenate $s$ and $a$ and learn a combined representation with a fully connected layer of 256 units before projecting it into Q-value (the output layer) for the state $s$ and given the action $a$ in $s$. This implementation of the Actor and Critic networks is inspired by a related open-source implementations available on the Web\footnote{github.com/yanpanlau/DDPG-Keras-Torcs}.

The network architecture for AP1 predictor is identical to that of Critic except that instead of Q-value, the combined representation of $s$ and $a$ is projected into two class (binary classification) output through a softmax projection (classification in this case) layer. 

\textbf{Hyper-parameter Settings.} The empirically chosen parameters are: learning rates for Actor is set as 0.0001, Critic as 0.001 and AP predictor as 0.001, the regularization parameter $\lambda$ as 0.01, $\delta_{acc}=0.7$, discount factor for Critic updates as 0.9, target network update parameter as 0.001, $\alpha_e$ as 0.5 and $\alpha_{tr}$ as 0.9, replay buffer size as 100k, knowledge buffer ($\mathcal{K}$) size as 10k (stores tuples in 9:1 ratio as training and validation examples), batch size as 128, sample size ($|\mathcal{A}_{s_t}|$) as 128 used for AP1-based guidance. $t_o$ is set as 200 and $t_e$ is set as 1200 for both 15k training and 3k training experiments. Sample size ($N_E$) for building the dataset for training AP1 predictor at each step is set as 2k and validation sample dataset size as 200 which is used to compute validation accuracy of AP1 predictor at each step of RL training. We employed the popularly used Ornstein-Uhlenbeck process for noise-based exploration with $\sigma = 0.3$ and $\theta = 0.15$ following standard settings for DDPG exploration. We train both networks with Adam optimizer.

\section{Flappy Bird: Description and Implementation Details} 

\textbf{Network Architecture.}  We use deep convolution network for constructing the double DQN (DDQN) following (Hasselt et al. 2015) and an existing open-source implementation\footnote{github.com/yenchenlin/DeepLearningFlappyBird}. The input to the DDQN network is a 80x80x4 tensor containing a rescaled, and gray-scale, version of the last four frames. The first convolution layer convolves the input with 32 filters of size 8 (stride 4), the second layer has 64 filters of size 4 (stride 2), the final convolution layer has 64 filters of size 3 (stride 1). In between the first and second convolution layer, we apply a max pooling layer of size 2 (stride 2) with 'SAME' padding. The representation obtained in the third convolution layer is flattened and fed to a fully-connected (FC) hidden layer of 512 units to get a representation (say, hidden representation $h_s$) which is then projected into Q-value (output layer) of size 2 (there are two possible actions for the Flappy bird game, flapping or not flapping).  

The architecture of the AP1 predictor is built as a shared network (shared weights) with that of DDQN upto the layer learning the state representation $h_s$. We use a FC layer of 256 units to learn representation of an action $a$. Then, we concatenate two representations (i.e., $h_s$ and representation of $a$) and learn a combined representation of the concatenated vector using another FC layer of 256 units. Finally, the combined representation is projected into two class (binary classification) output through a softmax projection.

Note that, due to the shared representation ($h_s$) learning of state $s$, the parameters of the shared network are trained with both AP1 predictor loss (equation 6, in the paper) and RL loss (equation 2, in the paper). The AP1 predictor loss being a supervised learning loss function with fixed target labels (unlike estimated target Q values) accelerates the training. For the lane keeping task, the network architecture being much simpler, we can train two networks (RL and AP1 predictor) quickly without any need for parameter sharing. 

Although DDQN-AP2 does not require an AP predictor, we observed performance improvement when we trained DDQN-AP2 parameters (upto layer $h_s$) with cross entropy loss [like in DDQN-(AP1+AP2)] over examples annotated by type-2 AP function apart from the RL loss. Note that, as DDQN-AP2 performs non-permissible actions with (1-$\alpha$) probability, we can use type-2 AP function to label the executed actions and populate a knowledge buffer just like in case of AP1 based guidance. The results reported in Table 2 (in the paper) corresponds to this version of DDQN-AP2.

\textbf{Hyper-parameter Settings.} For Flappy bird, the empirically chosen hyper-parameters are: learning rates for DDQN as 5e-6 and AP predictor as 0.0001, regularization parameter $\lambda$ as 0.01, $\delta_{acc}=0.95$, discount factor as 0.99, target network update parameter as 0.001, $\alpha_e$ as 0.3 and $\alpha_{tr}$ as 0.8, replay buffer size as 50k, knowledge buffer ($\mathcal{K}$) size as 25k (stores tuples in 9:1 ratio as training and validation examples), batch size for training as 128, sample size ($|\mathcal{A}_{s_t}|$) as 2 used for AP-based guidance (as there are two possible actions for Flappy bird), $t_o$ as 1000, $t_e$ as 30k for 30k exploration steps training experiments (see Table 2, in the paper), $t_e$ as 60k for 60k exploration steps training experiments, sample size ($N_E$) for building dataset for training AP1 predictor at each step as 2k, and validation sample dataset size as 200 which is used to compute validation accuracy of AP1 predictor at each step of the training. We train DDQN and AP1 predictor networks with Adam optimizer and Gradient Descent optimizer respectively. 

For training of DDQN and DDQN-AP, we use $\epsilon$-greedy strategy for the action space exploration. For 30k annealing steps training experiments, we set initial $\epsilon$ as 1.0, final $\epsilon$ as 0.01 and annealing steps as 30k with observation phase of 1k steps. For 60k annealing steps training experiments, we set annealing steps as 60k keeping all other parameters same as that for 30k.

\section{Cart-Pole Balancing: Description and Implementation Details}

The cart-pole balancing task is about learning to balance a pole attached by an un-actuated joint to a cart, which moves along a frictionless track (see Figure 8). The pendulum (pole) starts upright, and the goal is to prevent it from falling over by increasing and reducing the cart's velocity. 
The agent receives a reward of 1.0 for every step taken, including the termination step. The problem is considered as solved when the average reward acquired by the agent is greater than or equal to 195.0 over 100 consecutive trials. Action space is discrete, where the agent can apply a force of +1 (push right) or -1 (push left) to the cart to control its movement. 

The state of the agent is represented as a real-valued vector involving four variables: (1) \textit{Cart Position} (position of the cart along frictionless track, ranging from -2.4 to 2.4); (2) \textit{Cart Velocity} (velocity of the cart, ranging from -$\infty$ to $\infty$); (3) \textit{Pole Angle} (angle of the pole with regard to the vertical axis of oscillation denoted as Y and ranges from $\sim -\ang{41.8}$ to $\sim \ang{41.8}$) and (4) \textit{Pole velocity} (velocity of the pole at its tip, ranging from -$\infty$ to $\infty$). We denote the Pole angle and Pole velocity in a given state $s_t$ as $\theta_t$ and $v_t$ onwards. 

An episode is terminated if one of the following conditions is satisfied in a given state- \textbf{(1)} Pole Angle is beyond $\pm$~\ang{12}; \textbf{(2)} Cart Position is beyond $\pm~$2.4 (center of the cart reaches the edge of the display) and \textbf{(3)} Episode length is greater than 200.

\textbf{Network Architecture.}  For cart-pole balancing, the DDQN network is a two-layer feed forward (fully connected (FC)) network with 16 units in layer-1 and 32 units in layer-2 followed by the action projection (output) layer. Here, we denote the output of layer-2 as $hfc2$. AP1 predictor is built as a shared network (shared weights) with that of DDQN upto the layer learning the state representation $hfc2$. We use a FC layer of 32 units to learn representation of an action $a$. Then, we concatenate two representations (i.e., $hfc2$ and representation of $a$) and learn a combined representation of the concatenated vector using another FC layer of 32 units. Finally, the combined representation is projected into two class (binary classification) output through a softmax projection layer. 

\textbf{Hyper-parameter Settings.} For cart-pole, the empirically chosen hyper-parameters are: learning rates for DDQN as 0.0005 and AP1 predictor as 0.001, regularization parameter $\lambda$ as 0.01, $\delta_{acc}=0.9$, discount factor as 0.99, target network update parameter as 0.001, $\alpha_e$ as 0.3 and $\alpha_{tr}$ as 0.7, replay buffer size as 50k, knowledge buffer ($\mathcal{K}$) size as 25k (stores tuples in 9:1 ratio as training and validation examples), batch size for training as 128, sample size ($|\mathcal{A}_{s_t}|$) as 2 used for AP-based guidance (as there are two possible actions), $t_o$ as 100, $t_e$ as 2k, sample size ($N_E$) for building dataset for training AP1 predictor at each step as 2k, and validation sample dataset size as 200 which is used to compute validation accuracy of AP1 predictor at each step of the training. We train DDQN and AP1 predictor networks with Adam optimizer and Gradient Descent optimizer respectively. For training of DDQN and DDQN-AP1, we use the $\epsilon$-greedy strategy for the action space exploration, with initial $\epsilon$ as 1.0, final $\epsilon$ as 0.01 and annealing steps as 2k.

\section{Virtual Stopping: Flappy Bird and Cart-pole Balancing}
In the lane keeping task of autonomous driving, the (continuous-valued) reward function in equation 4 (Section 4, in the paper) ensures that any non-permissible action [labeled by the AP1 function in equation 5 (Section 4, in the paper)] will always have less reward than that for a permissible one in a given state\footnote{Note that the reward function monotonically decreases when the car moves away from the track center on either side.}. Thus, virtual stopping is not necessary for this case. We did not notice any significant difference in RL training speed up for the lane keeping task as compared to that using virtual stopping. This is because, as non-permissible experiences receive less reward than permissible ones, repeated training with non-permissible experiences (sampled from replay buffer) makes the RL model to converge faster.

However, unlike lane keeping, for Flappy bird and Cart-pole Balancing, often permissible and non-permissible actions (labeled by the AP functions) receive the same reward from the environment (e.g. 0.1 for Flappy bird and 1.0 for cart-pole). This is because, in the Flappy bird game, whenever the bird crosses a pipe, it gets 1.0 immediate reward; if it crashes (episode ends), it gets -1.0 immediate reward and otherwise, if it remains alive, it gets 0.1 always. Thus 1.0 and -1.0 are delayed rewards. Similarly, in cart-pole, the agent receives 1.0 reward irrespective of whether the pole is in safe zone or danger zone until the episode ends. These constant rewards to all actions until the episode end is equivalent to delayed reward. Virtual stopping is designed to deal with these situations because AP1 predictor and AP functions (including type 1 and type 2) can differentiate between permissive and non-permissible actions during training. Based on the two games, we give -1 reward to each non-permissible action at virtual stopping (with end of episode flag being `True' recorded in non-permissible experience tuple) to enable the system to learn in a short horizon to speed up training.

\begin{figure*}[t!]
    \centering
    \subfigure[\small Pixel Copter (PLE)]{
		\label{fig:first}%
		\includegraphics[height=3.5cm]{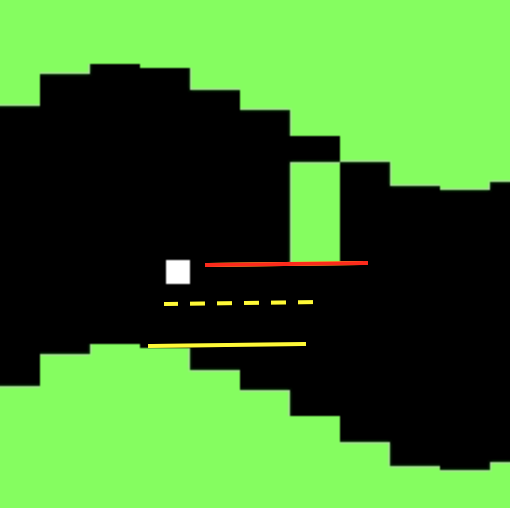}}%
	\quad
	\subfigure[\small Catcher (PLE)]{%
		\label{fig:third}%
		\includegraphics[height=3.5cm]{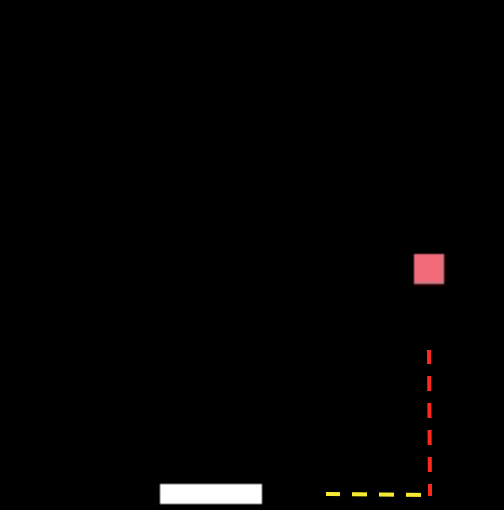}}
	\quad
	\subfigure[\small Pong (PLE)]{%
		\label{fig:third}%
		\includegraphics[height=3.5cm]{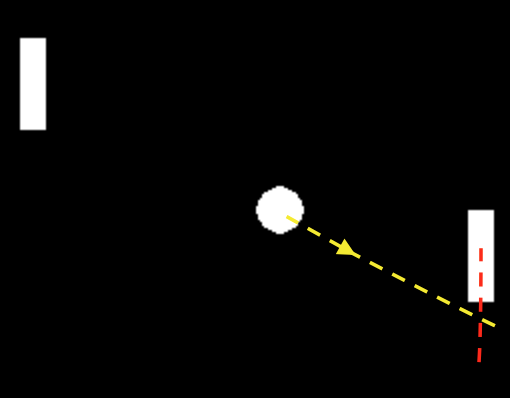}}
	\quad
	\subfigure[\small Tennis (Openai Gym)]{%
		\label{fig:third}%
		\includegraphics[height=3.5cm]{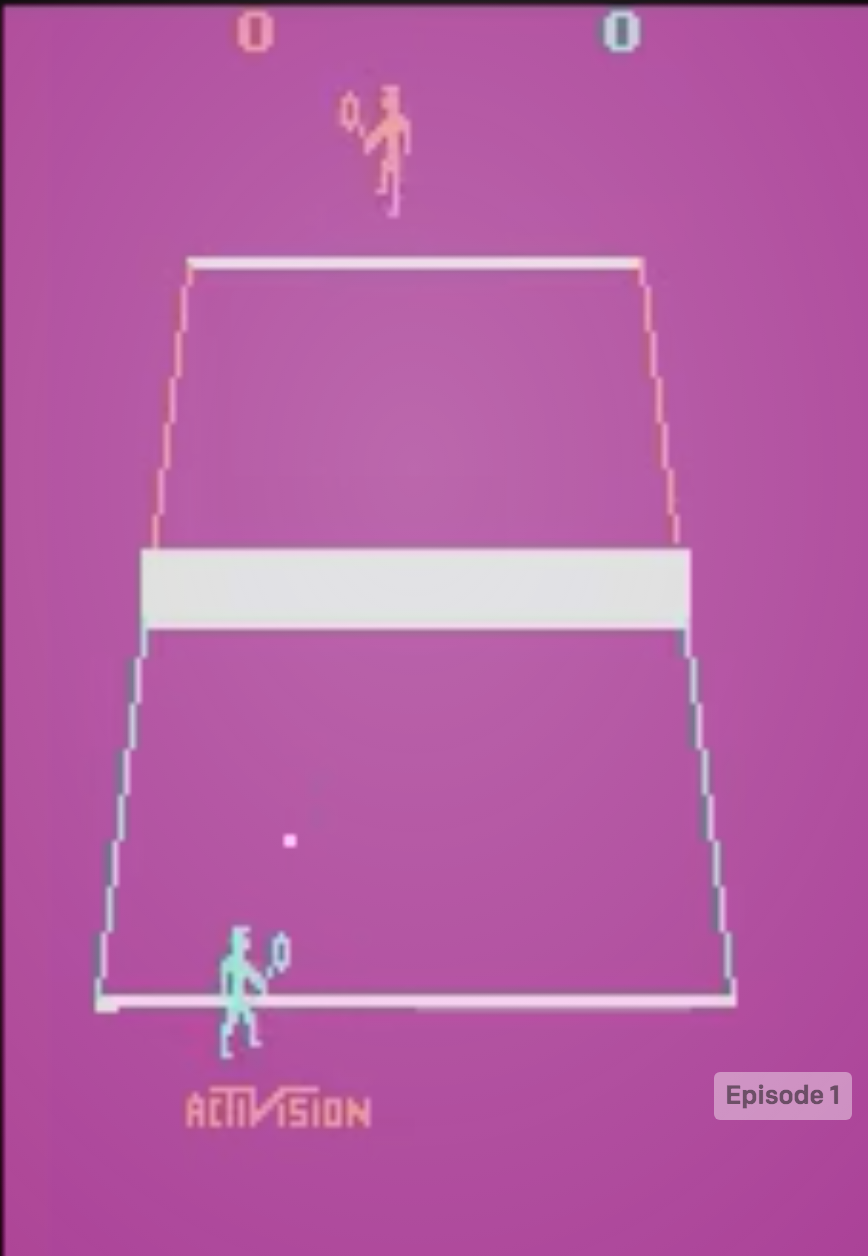}}
	\hspace{2cm}
    \quad
	\subfigure[\small Luner lander (Openai Gym)]{%
		\label{fig:third}%
		\includegraphics[height=3.5cm]{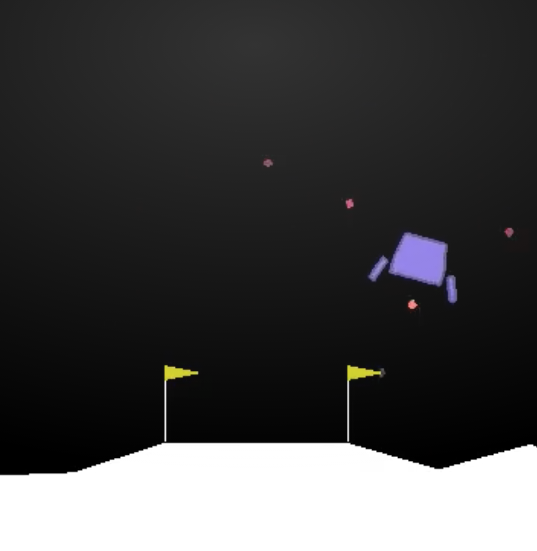}}
	\quad
	\subfigure[\small Reacher (Openai Gym)]{%
		\label{fig:third}%
		\includegraphics[height=3.5cm]{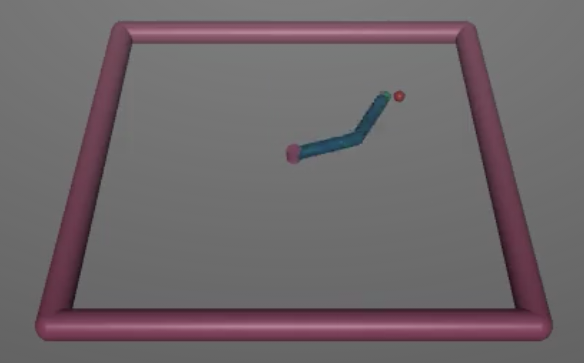}}
	\quad
	\subfigure[\small Turtlebot Navigation (Gym Gazebo)]{%
		\label{fig:third}%
		\includegraphics[height=3.5cm]{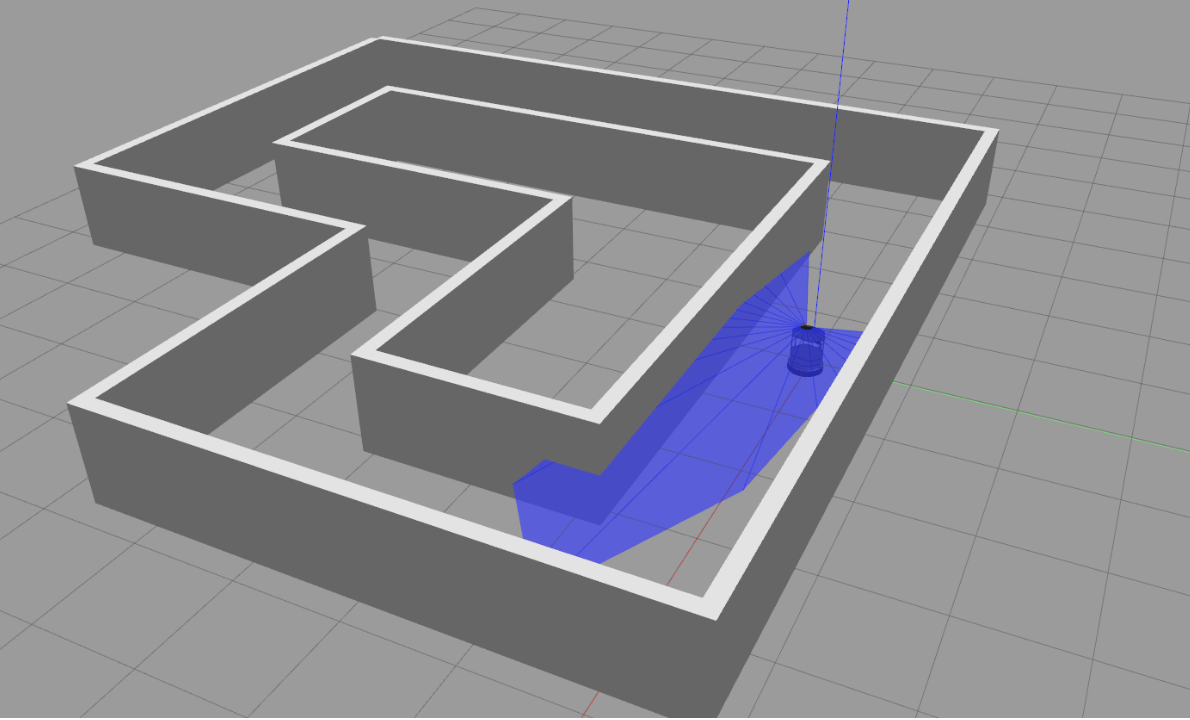}}
	\label{task_example}
	\vspace{-0.25cm}
	\caption{A (non-exhaustive) list of example RL problems with SAP property from  PyGame Learning Environment (PLE), Openai Gym and Gym Gazebo.}
	\vspace{-0.2cm}
\end{figure*}

\section{Hyper-parameter Tuning for AP-Guided Exploration}  
The parameters, $t_o$ and $t_e$ are adopted from existing RL methods, which denote the number of steps in the observation phase (during which epsilon is set as 1.0) and the number of steps in the exploration phase (during which epsilon is annealed from 1.0 to a low value, here 0.01) respectively.

The newly introduced parameters $\alpha_e$, $\alpha_{tr}$ and $\delta_{acc}$ have their distinct objectives as follows: $\alpha_e$ and $\alpha_{tr}$ are the values of $\alpha$ in exploration and post exploration (training) phase respectively and control the degree by which agent listens to the AP predictor. If we set $\alpha_e$ = 0, it denotes the agent does not utilize the permissibility knowledge at all during the exploration phase and so, the growth of the learning curve will be slow compared to that for $\alpha_e > 0$. Generally, we keep $0 < \alpha_e \leq 0.5$ to encourage more diverse exploration which also helps in AP1 predictor training and quickly populating the non-permissible replay buffer.  $\alpha_{tr}$ is generally set $1> \alpha_{tr} \geq 0.5$. High $\alpha_{tr}$ indicates the agent will more often listen to the AP1 predictor and thus, explore good actions more often than repeatedly executing bad actions (non-permissible ones).

$\delta_{acc}$ helps in measuring the reliability of the AP predictor and is set based on performance of the validation curve of the AP predictor. In particular, $\delta_{acc}$ sets an accuracy threshold on the performance of the AP predictor so that if the predictor’s validation accuracy is $< \delta_{acc}$, it indicates the learned model is not reliable and needs more training. If validation accuracy is $\geq \delta_{acc}$, then the learned model is considered reliable to assist the agent in exploration. Hence, after training for an initial number of steps, if the validation accuracy stays $\geq \delta_{acc}$, we can postpone the training of AP predictor until its validation accuracy at a given step falls below $\delta_{acc}$. In general, we set a fairly high threshold for $\delta_{acc}$ empirically that encourages both reliable consultation from the AP predictor in guided exploration process as well as periodic AP predictor retraining (to learn periodically with incoming new experiences).

\section{Applicability of SAP: More Examples}
SAP provides a framework for knowledge-driven reinforcement learning (RL). In particular, we show that domain knowledge can be encoded in AP functions which can be utilized by the RL agent for guided exploration to drastically reduce the RL training time. We have evaluated the proposed idea with three standard RL problems in the paper. However, there are a large number of other tasks or environments with the SAP property where it is not difficult to come up with a fairly good AP function based on the commonsense knowledge and understanding of the environment. Here, we discuss the AP function formulation strategy for 7 additional standard RL problems (see Fig. 12) from some popular RL platforms.  

\textbf{Pixel Copter} [Fig. 12(a)] is a side-scrolling game where the agent must successfully navigate through a cavern. This is a clone of the popular helicopter game but where the player is a humble pixel. If the pixel makes contact with anything green the game is over. So, a safe solution is to navigate through a path keeping the pixel as distant from the green cavern wall and objects as possible. We can formulate an AP function for this game borrowing ideas from the Flappy bird game. An example type-2 AP function would be: if the copter pixel at a given time step is above the \textit{upper boundary line} [horizontal line containing the bottom-most green pixel of upper green region, shown by the \textit{red solid line} in Fig. 12(a)], then the copter should not accelerate upwards further. Similarly, when the copter is below the \textit{lower boundary line} [horizontal line containing top-most green pixel of lower green region, shown by the \textit{yellow solid line} in Fig. 12(a)], it should not fall down further. Also, any upward (downward) movement that causes the copter to touch green pixel in upper (lower) region is also non-permissible [type-1 permissibility].

In \textbf{Catcher} [Fig. 12(b)] the agent must catch the falling fruit with its paddle, where the allowed actions are left and right movements to control the direction of the paddle. The paddle has a little velocity added to it to allow smooth movements. The game is over when the agent lose the number of lives set by init\_lives parameter. Here the catcher moves along a horizontal line. So, an optimal strategy to play the game is to make the catcher to reach the goal point [point of intersection of dotted red line and dotted yellow line shown in Fig. 12(b)] before the fruit (red object) reaches the goal point, otherwise the catcher would miss the fruit. So, a straightforward type 2 AP function for the task is, that if the fruit is falling on the left side of catcher (more precisely, with regard to the vertical line containing catcher's mid point), any rightward movement of the catcher is non-permissible and vice-versa. 

Fig. 12(c) shows the \textbf{pong game} that simulates 2D table tennis. The agent controls an in-game paddle which is used to hit the ball back to the other side. The agent controls the left paddle while the CPU controls the right paddle. The game is over if either the agent or CPU reaches the number of points set by MAX\_SCORE. A type 2 AP function for Pong can be formulated as follows: If the ball is moving towards the agent and away from the CPU and the \textit{projected goal point} (the intersection of ball's trajectory and vertical line of movement of the RL agent, shown as the point of intersection of the yellow and red dotted line in Fig. 12(c) respectively) is below the agent's current position (any part of the agent is not on projected goal point), any upward movement or no movement is non-permissible and vice versa. When the ball is moving away from the agent and towards the CPU, any action (upward or downward) taken by the agent is permissible. This is similar to how we apply our intuition while learning to play table tennis. An non-expert payer don't move randomly to hit the ball (like random exploration in RL), rather he/she tries to guess the ball's trajectory and then, attempts to move closer to ball before it reaches him/her (by applying prior knowledge and understanding the dynamics of the environment). Similar AP function can be formulated to learn to play the \textbf{Tennis} game [Fig. 12(d)].

Fig. 12(e) shows the \textbf{lunar lander} task from open-ai gym, where the agent must learn to successfully land a rocket on the landing pad on the surface of the Moon. The landing pad is always at a fixed coordinate (0, 0). Reward for moving from the top of the screen to landing pad and zero speed is about 100..140 points. If lander moves away from landing pad it loses reward back. Episode finishes if the lander crashes or comes to rest, receiving additional -100 or +100 points, respectively. Each leg ground contact is +10. Firing main engine is -0.3 points each frame. Solved is 200 points. Landing outside the landing pad is possible. Fuel is infinite, so an agent can learn to fly and then land on its first attempt. Four discrete actions are available: do nothing, fire left orientation engine, fire main engine, fire right orientation engine. An type-1 AP function for the task can be formulated as follows: If the agent moves away from the landing pad due to the execution of an action, the action is a non permissible one. If the orientation of the agent gets worse due to an action, (e.g., the agent's legs are not horizontally aligned and are more tilted in the new state than it was before), it will not lead to successful landing and so, the executed action increasing tilt is also non-permissible. Otherwise, the action is permissible. 

Fig. 12(f) shows the robot arm \textbf{Reacher} environment, where the agent must learn to move its arm to make the arm tip reach a goal point (or within a pre-defined circular neighborhood around the goal point). For this task, any robot arm movement that causes the robot arm to be away from the goal point (increases the distance between the robot arm tip and the goal point), is non-permissible. Otherwise, the action is permissible. 

Fig. 12(g) shows a \textbf{Turtlebot navigation} environment, where the agent (Turtlebot) must learn to navigate autonomously through a maze without hitting the wall. This problem is similar to the steering control problem in autonomous driving. 

Besides the aforementioned example tasks or environments, many other RL problems can be found in practice where SAP is applicable and useful for RL training. In our current work, we only dealt with one dimensional discrete and continuous space RL problems. However, the concept of SAP is also applicable and can be further extended to environments with multi-dimensional discrete and continuous action spaces. For example, considering autonomous driving, we can define three AP functions independently, one for steering control (as we did in our work) and other two for speed control like for break and acceleration. Their interaction will be quite interesting. We feel it will improve RL learning even further because the reduction in each dimension will result in much more reduction in the cross product. We leave the formulation of SAP for this multi-dimensional action space case as our future work.  

\section{Does SAP counter RL's philosophy?}
It is important to stress that \textbf{choosing a permissible action is by no means a greedy decision that counters RL's philosophy of sacrificing the immediate reward} to optimize for the accumulated reward over the long run. The purpose of AP function is only to cut off exploration space for a given state, and to enable RL to not explore non-permissible actions in similar states again and again. In particular, it estimates a permissible action space in a given state and prioritize exploration of those actions in that state compared to the non-permissible ones. There may be multiple permissible actions in a given state to choose from. But, SAP does not tell you which one is optimal at that point. Rather, SAP only tells you which one you should definitely avoid exploring, as there is a better option (action) available in that state to explore. And, it is the RL algorithm’s job to find out the optimal policy from the permissible action space in the long run. Thus, we are not chopping off any optimal solution in AP-based guidance [note, even the RL agent explores non-permissible actions with ($1- \alpha$) probability, see the paper]. The idea of SAP is not contradictory to RL.

It is hard to imagine that humans learn like a typical pure RL algorithm. We humans do not normally blindly try all possible actions. Our prior knowledge about the environment often can tell us what movements/actions will not help us reach our goals. SAP guidance aims to achieve the same goal through human-AI collaboration.

\end{document}